\documentclass[11pt]{article}

\usepackage{jair}
\usepackage{theapa}

\usepackage{times,pslatex,amsmath,tabularx,float,dcolumn}

\usepackage{graphicx,url}

\usepackage{examples}

\usepackage{macros,markup,epsfig}

\newcommand{\argmax}{\operatornamewithlimits{argmax}}

\newcommand{\shortcitegen}[1]{\shortciteauthor{#1}'s \citeyear{#1}}

\newcolumntype{d}{D{.}{.}{-1}}  
\newcolumntype{t}{D{.}{.}{1}}   
\newcolumntype{f}{D{.}{.}{6}}   

\newcommand{\labe}[1]{\multicolumn{1}{c|}{#1}}

\jairheading{27}{2006}{85--117}{01/06}{09/06}
\ShortHeadings{Learning Sentence-internal Temporal Relations}{Lapata \& Lascarides}
\firstpageno{85}

\title{Learning Sentence-internal Temporal Relations}
\author{\name Mirella Lapata \email mlap@inf.ed.ac.uk\\
\name Alex Lascarides \email alex@inf.ed.ac.uk\\
\addr School of Informatics,\\
University of Edinburgh,\\
2 Buccleuch Place,\\
Edinburgh, EH8 9LW,\\
Scotland, UK}

\begin{document}

\maketitle

\begin{abstract}
  In this paper we propose a data intensive approach for inferring
  sentence-internal temporal relations. Temporal inference is relevant
  for practical NLP applications which either extract or synthesize
  temporal information (e.g.,~summarisation, question answering).  Our
  method bypasses the need for manual coding by exploiting the
  presence of markers like \word{after}, which overtly signal a
  temporal relation. We first show that models trained on main and
  subordinate clauses connected with a temporal marker achieve good
  performance on a pseudo-disambiguation task simulating temporal
  inference (during testing the temporal marker is treated as unseen
  and the models must select the right marker from a set of possible
  candidates).  Secondly, we assess whether the proposed approach
  holds promise for the semi-automatic creation of temporal
  annotations.  Specifically, we use a model trained on noisy and
  approximate data (i.e.,~main and subordinate clauses) to predict
  intra-sentential relations present in TimeBank, a corpus annotated
  rich temporal information.  Our experiments compare and contrast
  several probabilistic models differing in their feature space,
  linguistic assumptions and data requirements.  We evaluate
  performance against gold standard corpora and also against human
  subjects. 
\end{abstract}

\section{Introduction}
\label{sec:introduction}

The computational treatment of temporal information has recently
attracted much attention, in part because of its increasing importance
for potential applications.  In multidocument summarization, for
example, information that is to be included in the summary must be
extracted from various documents and synthesized into a meaningful
text.  Knowledge about the temporal order of events is important for
determining what content should be communicated and for correctly
merging and presenting information in the summary.  Indeed, ignoring
temporal relations in either the information extraction or the summary
generation phase may result in a summary which is misleading with
respect to the temporal information in the original documents.  In
question answering, one often seeks information about the temporal
properties of events (e.g.,~\word{When did X resign?}) or how events
relate to each other (e.g.,~\word{Did X resign before Y?}).

An important first step towards the automatic handling of temporal
phenomena is the analysis and identification of time expressions. Such
expressions include absolute date or time specifications
(e.g.,~\word{October 19th, 2000}), descriptions of intervals
(e.g.,~\word{thirty} \word{years}), indexical expressions
(e.g.,~\word{last week}), etc. It is therefore not surprising that
much previous work has focused on the recognition, interpretation, and
normalization of time expressions\footnote{See also the Time
  Expression Recognition and Normalisation (TERN) evaluation exercise
  (\url{http://timex2.mitre.org/tern.html}).}
\cite{Wilson:ea:01,Schilder:Habel:01,Wiebe:ea:98}. Reasoning with
time, however, goes beyond temporal expressions; it also involves
drawing inferences about the temporal relations among events and other
temporal elements in discourse.  An additional challenge to this task
stems from the nature of temporal information itself, which is often
implicit (i.e.,~not overtly verbalized) and must be inferred using
both linguistic and non-linguistic knowledge.

Consider the examples in~\exr{ex:temp} taken from
\citeA{Katz:Arosio:01}. Native speakers can infer that John first
met and then kissed the girl; that he left the party after kissing the
girl and then walked home; and that the events of talking to her and
asking her for her name temporally overlap (and occurred before he
left the party).

\ex{
\item \label{ex:temp}
\ex{
\item \label{ex:temp:1} John kissed the girl he met at a party. 
\item \label{ex:temp:2} Leaving the party, John walked home.
\item \label{ex:temp:3} He remembered talking to her and asking her for her name. 
}}

The temporal relations just described are part of the interpretation
of this text, even though there are no overt markers, such as
\word{after} or \word{while}, signaling them.  They are inferable from
a variety of cues, including the order of the clauses, their
compositional semantics (e.g.,~information about tense and aspect),
lexical semantics and world knowledge. In this paper we describe a
data intensive approach that automatically captures information
pertaining to the temporal relations among events like the ones
illustrated in~\exr{ex:temp}.

A standard approach to this task would be to acquire a model of
temporal relations from a corpus annotated with temporal information.
Although efforts are underway to develop treebanks marked with
temporal relations \cite{Katz:Arosio:01} and devise annotation schemes
that are suitable for coding temporal relations
\cite{Sauri:ea:04,Ferro:ea:00,Setzer:Gaizauskas:01}, the existing
corpora are too small in size to be amenable to supervised machine
learning techniques which normally require thousands of training
examples.  The TimeBank\footnote{Available from
  \url{http://www.cs.brandeis.edu/~jamesp/arda/time/timebank.html}}
corpus, for example, contains a set of 186 news report documents
annotated with the TimeML mark-up language for temporal events and
expressions (for details, see Sections~\ref{sec:related-work}
and~\ref{sec:timebank-experiment}).  The corpus consists of 68.5K
words in total. Contrast this with the Penn Treebank, a corpus which
is often used in many NLP tasks and contains approximately 1M words
(i.e.,~it is 16~times larger than TimeBank).  The annotation of
temporal information is not only time-consuming but also error prone.
In particular, if there are $n$ kinds of temporal relations, then the
number of possible relations to annotate is a polynomial of factor $n$
on the number of events in the text.  \citeA{pustejovsky:etal:2003}
found evidence that this annotation task is sufficiently complex that
human annotators can realistically identify only a small number of the
temporal relations in text, thus compromising recall.

In default of large volumes of data labeled with temporal information,
we turn to unannotated texts which nevertheless contain expressions
that overtly convey the information we want our models to learn.
Although temporal relations are often underspecified, sometimes there
are temporal markers, such as \word{before}, \word{after}, and
\word{while}, which make relations among events explicit:

\ex{
\item \label{ex:con}
\ex{
\item Leonard Shane, 65 years old, held the post of president before
  William Shane, 37, was elected to it last year.
\item The results were announced after the market closed. 
\item Investors in most markets sat out while awaiting the U.S. trade figures.
}}

It is precisely this type of data that we will exploit for making
predictions about the temporal relationships among events in text.  We
will assess the feasibility of such an approach by initially focusing
on sentence-internal temporal relations. From a large corpus, we will
obtain sentences like the ones shown in~\exr{ex:con}, where a main
clause is connected to a subordinate clause with a temporal marker,
and we will develop a probabilistic framework where the temporal
relations will be inferred by gathering informative features from the
two clauses.  Our models will view the marker from each sentence in
the training corpus as the label to be learned. In the test corpus the
marker will be removed and the models' task will be to pick the most
likely label---or equivalently marker.

We will also examine whether models trained on data containing main
and subordinate clauses together with their temporal connectives can
be used to infer relations among events when temporal information is
underspecified and overt temporal markers are absent (as in each of
the three sentences in~(\ref{ex:temp})). For this purpose, we will
resort to the TimeBank corpus. The latter contains detailed
annotations of events and their temporal relations irrespectively of
whether connectives are present or not.  Using the TimeBank
annotations solely as test data, we will assess whether the approach
put forward generalizes to different structures and corpora. Our
evaluation study will also highlight whether a model learned from
unannotated examples could alleviate the data acquisition bottleneck
involved in the creation of temporal annotations. For example, by
automatically creating a high volume of annotations which could be
subsequently corrected manually.

In attempting to infer temporal relations probabilistically, we
consider several classes of models with varying degrees of
faithfulness to linguistic theory. Our models differ along two
dimensions: the employed feature space and the underlying independence
assumptions. We compare and contrast models which utilize
word-co-occurrences with models which exploit linguistically motivated
features (such as verb classes, argument relations, and so on).
Linguistic features typically allow our models to form generalizations
over {\em classes} of words, thereby requiring less training data than
word co-occurrence models.  We also compare and contrast two kinds of
models: one assumes that the properties of the two clauses are
mutually independent; the other makes slightly more realistic
assumptions about dependence. (Details of the models and features used
are given in Sections~\ref{sec:model}
and~\ref{sec:parameter-estimation}).  We furthermore explore the
benefits of ensemble learning methods for the temporal interpretation
task and show that improved performance can be achieved when different
learners (modeling sufficiently distinct knowledge sources) are
combined.  Our machine learning experiments are complemented by a
study in which we investigate human performance on the interpretation
task thereby assessing its feasibility and providing a ceiling on
model performance.

The next section gives an overview of previous work in the area of
computing temporal information and discusses related work which
utilizes overt markers as a means for avoiding manual labeling of
training data.  Section~\ref{sec:model} describes our probabilistic
models and Section~\ref{sec:parameter-estimation} discusses our
features and the motivation behind their selection. Our experiments
are presented in
Sections~\ref{sec:exper-1:-conf}--\ref{sec:timebank-experiment}.
Section~\ref{sec:discussion} offers some discussion and concluding
remarks.


\section{Related Work}
\label{sec:related-work}

Traditionally, methods for inferring temporal relations among events
in discourse have utilized a semantics and inference-based approach.
This involves complex reasoning over a variety of rich information
sources, including elaborate domain knowledge and detailed logical
form representations
\shortcite<e.g.,>{dowty:1986,hwang:schubert:1992,hobbs:etal:1993,lascarides:asher:1993a,kamp:reyle:1993,kehler:2002}.
This approach, while theoretically elegant, is impractical except for
applications in very narrow domains.  This is for (at least) two
reasons.  First, grammars that produce detailed semantic
representations inevitably lack linguistic coverage and are brittle in
the face of natural data; similarly, the representations of domain
knowledge can lack coverage.  Secondly, the complex reasoning required
with these rich information sources typically involves nonmonotonic
inferences \shortcite<e.g.,>{hobbs:etal:1993,lascarides:asher:1993a},
which become intractable except for toy examples.

\citeA{allen:1995}, \citeA{hitzeman:etal:1995}, and
\citeA{han:lavie:2004} propose more computationally tractable
approaches to infer temporal information from text, by hand-crafting
algorithms which integrate shallow versions of the knowledge sources
that are exploited in the above theoretical literature
\shortcite<e.g.,>{hobbs:etal:1993,kamp:reyle:1993}.  While this type
of symbolic approach is promising, and overcomes some of the
impracticalities of utilizing full logical forms and complex reasoning
over rich domain knowledge sources, it is not grounded in empirical
evidence of the way the various linguistic features contribute to the
temporal semantics of a discourse; nor are these algorithms evaluated
against real data.  Moreover, the approach is typically
domain-dependent and robustness is compromised when porting to new
domains or applications.

Acquiring a model of temporal relations via machine learning over a
training corpus promises to provide systems which are precise, robust,
and grounded in empirical evidence. A number of markup languages have
recently emerged that can greatly facilitate annotation efforts in
creating suitable corpora. A notable example is TimeML
(\citeR{pustejovsky:etal:2004}; see also the annotation scheme by
\citeR{Katz:Arosio:01}), a metadata standard for expressing
information about the temporal properties of events and temporal
relations between them.  The scheme can be used to annotate a variety
of temporal expressions, including tensed verbs, adjectives and
nominals that correspond to times, events or states.  The type of
temporal information that can be expressed on these various linguistic
expressions includes the class of event, its tense, grammatical
aspect, polarity (positive or negative), the time denoted (e.g.,~one
can annotate \word{yesterday} as denoting the day before the document
date), and temporal relations between pairs of eventualities and
between events and times. TimeML's expressive capabilities are
illustrated in the TimeBank corpus which contains temporal annotations
of news report documents (for details, see
Section~\ref{sec:timebank-experiment}).

\citeA{Mani:ea:03} and \citeA{mani:schiffman:2005}
demonstrate that TimeML-compliant annotations are useful for learning
a model of temporal relations in news text. They focus on the problem
of ordering pairs of successively described events. A decision tree
classifier is trained on a corpus of temporal relations provided by
human subjects.  Using features such as the position of the sentence
within the paragraph (and the position of the paragraph in the text),
discourse connectives, temporal prepositions and other temporal
modifiers, tense features, aspect shifts and tense shifts, their best
model achieves 75.4\% accuracy in identifying the temporal order of
events.  \citeA{Boguraev:Ando:05} use semi-supervised learning
for recognizing events and inferring temporal relations (between an
event and a time expression). Their method exploits TimeML annotations
from the TimeBank corpus and large amounts of unannotated data. They
first build a classifier from the TimeML annotations using a variety
of features based on syntactic analysis and the identification of
temporal expressions. The original feature vectors are next augmented
with unlabeled data sharing structural similarities with the training
data. Their algorithm yields performances well above the baseline for
both tasks.

Conceivably, existing corpus data annotated with {\em discourse
  structure}, such as the {\sc rst} treebank
\shortcite{carlson:etal:2001}, might be reused to train a temporal
relations classifier.  For instance, for text spans connected with
{\sc result}, it is implied by the semantics of this relation that the
events in the first span temporally precede the second; thus, a
classifier of rhetorical relations could indirectly contribute to a
classifier of temporal relations.  Corpus-based methods for computing
discourse structure are beginning to emerge
\shortcite<e.g.,>{marcu:1999,Soricut:Marcu:03,baldridge:lascarides:2005a}.
But there is currently no automatic mapping from discourse structures
to their temporal consequences; so although there is potential for
eventually using linguistic resources labeled with discourse
structure to acquire a model of temporal relations, that potential
cannot be presently realized.
 
Continuing on the topic of discourse relations, it is worth mentioning
\citeA{Marcu:Echihabi:02} whose approach bypasses altogether the need
for manual coding in a supervised learning setting. A key insight in
their work is that rhetorical relations (e.g.,~{\sc explanation} and
{\sc contrast}) are sometimes signaled by a discourse connective
(e.g.,~\word{because} for {\sc explanation} and \word{but} for {\sc
  contrast}).  They extract sentences containing such markers from a
corpus, and then (automatically) identify the text spans connected by
the marker, remove the marker and replace it with the rhetorical
relation it signals.  A Naive Bayes classifier is trained on this
automatically labeled data.  The model is designed to be maximally
simple and employs solely word bigrams as features.  Specifically,
bigrams are constructed over the cartesian product of words occurring
in the two text spans and it is assumed that word pairs are
conditionally independent. Marcu and Echihabi demonstrate that such a
knowledge-lean approach performs well, achieving an accuracy of
49.70\% when distinguishing six relations (over a baseline
of~16.67\%).  However, since the model relies exlusively on
word-co-occurrences, an extremely large training corpus (in the order
of 40 M sentences) is required to avoid sparse data (see
\citeR{sporleder:lascarides:2005a} for more detailed discussion
on the tradeoff between training size and feature space for
discourse-based models).

In a sense, when considering the complexity of various models used to
infer temporal and discourse relations, Marcu and Echihabi's (2002)
model lies at the simple extreme of the spectrum, whereas the
semantics and inference-based approaches to discourse interpretation
\shortcite<e.g.,>{hobbs:etal:1993,Asher:Lascarides:03} lie at the
other extreme, for these latter theories assume \emph{no independence}
among the properties of the spans, and they exploit linguistic and
non-linguistic features to the full.  In this paper, we aim to explore
a number of probabilistic models which lie in between these two
extremes, thereby giving us the opportunity to study the tradeoff
between the complexity of the model on the one hand, and the amount of
training data required on the other.  We are particularly interested
in assessing the performance of models on smaller training sets than
those used by \citeA{Marcu:Echihabi:02}; such models will be useful
for classifiers that are trained on data sets where relatively rare
temporal markers are exploited.
  
Our work differs from \citeA{Mani:ea:03} and
\citeA{Boguraev:Ando:05} in that we do not exploit manual
annotations in any way. Our aim is however similar, since we also
infer temporal relations between pairs of events. We share with
\citeA{Marcu:Echihabi:02} the use of data with overt markers as a
proxy for hand coded relations. Apart from the fact that our
interpretation task is different from theirs, our work departs from
\citeA{Marcu:Echihabi:02} in three further important ways. First,
we propose alternative models and explore the contribution of
linguistic information to the inference task, investigating how this
enables one to train on considerably smaller data sets.  Secondly, the
proposed models are used to infer relations between events in a more
realistic setting, where temporal markers are naturally absent
(i.e.,~the test data is not simulated by removing the markers in
question).  And finally, we evaluate the models against human subjects
performing the same task, as well as against a gold standard corpus.


\section{Problem Formulation}
\label{sec:model}

Given a main clause and a subordinate clause attached to it, our task
is to infer the temporal marker linking the two clauses.
$P(S_{M},t_{j},S_{S})$ represents the probability that a marker
$t_{j}$ relates a main clause $S_{M}$ and a subordinate clause
$S_{S}$. We aim to identify which marker $t_j$ in the set of possible
markers $T$ maximizes the joint probability $P(S_{M},t_{j},S_{S})$:

\begin{equation}
\label{eq:1} 
\begin{array}[t]{lll}
t^{*} & = &\argmax\limits_{t_j \in T} P(S_{M}, t_j, S_{S}) \\
t^{*} & = &\argmax\limits_{t_j \in T} P(S_M) P(S_{S}|S_M) P(t_j | S_{M},S_{S}) \\
\end{array}
\end{equation}
We ignore the terms $P(S_{M})$ and $P(S_{S}|S_M)$ in \exr{eq:1} as
they are constant.  We use Bayes' Rule to calculate $P(t_j |
S_{M},S_{S})$:

\begin{equation}
\label{eq:2} 
\begin{array}[t]{lll}
t^{*} & = &\argmax\limits_{t_j \in T} P(t_j | S_{M},S_{S}) \\
t^{*} & = &\argmax\limits_{t_j \in T} P(t_j) P(S_{M},S_{S}|t_j)\\
t^{*} & = &\argmax\limits_{t_j \in T} P(t_j) P(a_{\langle M,1 \rangle}
\cdots a_{\langle S,n \rangle} | t_j)
\end{array}
\end{equation}
$S_{M}$ and $S_{S}$ are vectors of features $a_{\langle M,1 \rangle}
\cdots a_{\langle M,n \rangle}$ and $a_{\langle S,1 \rangle} \cdots
a_{\langle S,n \rangle}$ characteristic of the propositions occurring
with the marker $t_{j}$ (our features are described in detail in
Section~\ref{sec:feat-gener-temp}). Estimating the different
$P(a_{\langle M,1 \rangle} \cdots a_{\langle S,n \rangle} | t_j)$
terms will not be feasible unless we have a very large set of training
data. We will therefore make the simplifying assumption that a
temporal marker $t_j$ can be determined by observing feature {\em
  pairs} representative of a main and a subordinate clause. We further
assume that these feature pairs are conditionally independent given
the temporal marker and are not arbitrary: rather than considering all
pairs in the cartesian product of $a_{\langle M,1 \rangle} \cdots
a_{\langle S,n \rangle}$ \shortcite<see>{Marcu:Echihabi:02}, we
restrict ourselves to feature pairs that belong to the same class~$i$.
Thus, the probability of observing the conjunction $a_{\langle M,1
  \rangle} \cdots a_{\langle S,n \rangle}$ given $t_j$ is:
\begin{equation}
\label{eq:3}
t^{*} =  \argmax\limits_{t_j \in T}  P(t_j) \prod\limits_{i=1}^{n}
\biggl( P(a_{\langle M,i, \rangle}, a_{\langle S,i
\rangle}|t_{j})  \biggr) 
\end{equation}
For example, if we were assuming our feature space consisted solely of
nouns and verbs, we would estimate $P(a_{\langle M,i, \rangle},
a_{\langle S,i \rangle}|t_{j})$ by taking into account all noun-noun
and verb-verb bigrams that are attested in~$S_{S}$ and $S_{M}$ and
co-occur with~$t_{j}$.

The model in~(\ref{eq:2}) can be further simplified by assuming that
the likelihood of the subordinate clause $S_{S}$ is conditionally
independent of the main clause $S_{M}$ (i.e.,~$P(S_{S},S_{M}|t_j)
\approx P(S_{S}|t_j)P(S_{M}|t_j)$).  The assumption is clearly a
simplification but makes the estimation of the probabilities
$P(S_{M}|t_j)$ and $P(S_{S}|t_j)$ more reliable in the face of sparse
data:
\begin{equation}
\label{eq:4} 
\begin{array}[t]{lll}
t^{*} & \approx & \argmax\limits_{t_j \in T}  P(t_j) P(S_{M}|t_j) P(S_{S}|t_j)
      
\end{array}
\end{equation}
$S_{M}$ and $S_{S}$ are again vectors of features $a_{\langle M,1
  \rangle} \cdots a_{\langle M,n \rangle}$ and $a_{\langle S,1
  \rangle} \cdots a_{\langle S,n \rangle}$ representing the clauses
co-occurring with the marker $t_{j}$. Now individual features (instead
of feature pairs) are assumed to be conditionally independent given
the temporal marker, and therefore:
\begin{equation}
\label{eq:5}
t^{*} =  \argmax\limits_{t_j \in T}  P(t_j) \prod\limits_{i=1}^{n}
\biggl( P(a_{\langle M,i \rangle}|t_{j}) P(a_{\langle S,i
\rangle}|t_{j}) \biggr) 
\end{equation}
Returning to our example feature space of nouns and verbs,
$P(a_{\langle M,i \rangle}|t_{j})$ and $P(a_{\langle S,i
  \rangle}|t_{j})$ will be now estimated by considering how often
verbs and nouns co-occur with $t_{j}$. These co-occurrences will be
estimated separately for main and subordinate clauses.

Throughout this paper we will use the terms \emph{conjunctive} for
model~(\ref{eq:3}) and \emph{disjunctive} for model~(\ref{eq:5}).  We
effectively treat the temporal interpretation problem as a
disambiguation task. From a (confusion) set~$T$ of temporal markers,
e.g.,~\{\word{after, before, since}\}, we select the one that
maximizes~\exr{eq:3} or~\exr{eq:5} (see
Section~\ref{sec:parameter-estimation} for details on our confusion
set and corpus). The conjunctive model explicitly captures
dependencies between the main and subordinate clauses, whereas the
disjunctive model is somewhat simplistic in that relationships between
features across the two clauses are not represented directly. However,
if two values of these features for the main and subordinate clauses
co-occur frequently with a particular marker, then the conditional
probability of these features on that marker will approximate the
right biases.

The conjunctive model is more closely related to the kinds of symbolic
rules for inferring temporal relations that are used in semantics and
inference-based accounts \shortcite<e.g.,>{hobbs:etal:1993}. Many
rules typically draw on the {\em relationships} between the verbs in
both clauses, or the nouns in both clauses, and so on.  Both the
disjunctive and conjunctive models are different from
\shortcitegen{Marcu:Echihabi:02} model in several respects. They
utilize linguistic features rather than word bigrams.  The conjunctive
model's features are two-dimensional with each dimension belonging to
the same feature class. The disjunctive model has the added difference
that it assumes independence in the features attested in the two
clauses.

%
%



\section{Parameter Estimation}
\label{sec:parameter-estimation}

We can estimate the parameters for our models from a large corpus. In
their simplest form, the features $a_{\langle M,i \rangle}$ and
$a_{\langle S,i \rangle}$ can be the words making up main and
subordinate clauses. In order to extract relevant features, we first
identify clauses in a hypotactic relation, i.e.,~main clauses of which
the subordinate clause is a constituent. In the training phase, we
estimate the probabilities $P(a_{\langle M,i \rangle}|t_{j})$ and
$P(a_{\langle S,i \rangle}|t_{j})$ for the disjunctive model by simply
counting the occurrence of the features $a_{\langle M,i \rangle}$ and
$a_{\langle S,i \rangle}$ with marker~$t_{j}$ (i.e.,~$f(a_{\langle M,i
  \rangle},t_j)$) and ($f(a_{\langle S,i \rangle},t_j)$).  In essence,
we assume for this model that the corpus is representative of the way
various temporal markers are used in English.  For the conjunctive
model we estimate the co-occurrence frequencies $f(a_{\langle M,i
  \rangle}, a_{\langle S,i \rangle},t_{j})$. Features with zero counts
are smoothed in both models; we adopt the $m$-estimate with uniform
priors, with $m$ equal to the size of the feature space
\shortcite{Cestnik:90}. 


 \begin{figure}[t!]
 \vspace{6ex}
\includegraphics[width=0.0035\textwidth]{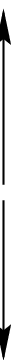} 
 \end{figure}

\begin{figure}[t!]
\vspace{-24ex}
\begin{small}
\begin{verbatim}
(S1 (S (NP (DT The) (NN company))
     (VP (VBD said)
       (S (NP (NNS employees))
         (VP (MD will)
          (VP (VB lose)
           (NP (PRP their) (NNS jobs))
           (SBAR-TMP (IN after)
            (S (NP (DT the) (NN sale))
             (VP (AUX is) (VP (VBN completed)))
  ))))))))
\end{verbatim}
\end{small}
\caption{\label{fig:tree-extract:a} Extraction of main and subordinate
     clause from parse tree} 
\end{figure}

\subsection{Data Extraction}
\label{sec:data-extraction}

In order to obtain training and testing data for the models described
in the previous section, subordinate clauses (and their main clause
counterparts) were extracted from the {\sc Bllip} corpus (30~M words).
The latter is a Treebank-style, machine-parsed version of the Wall
Street Journal (WSJ, years 1987--89) which was produced using
\shortcitegen{charniak:2000} parser.  Our study focused on the
following (confusion) set of temporal markers: \{\word{after, before,
  while, when, as, once, until, since}\}. We initially compiled a list
of all temporal markers discussed in \citeA{Quirk:ea:85} and
eliminated markers with frequency less than 10 per million in our
corpus.

We extract main and subordinate clauses connected by temporal
discourse markers, by first traversing the tree top-down until we
identify the tree node bearing the subordinate clause label we are
interested in and then extract the subtree it dominates.  Assuming we
want to extract \word{after} subordinate clauses, this would be the
subtree dominated by SBAR-TMP in Figure~\ref{fig:tree-extract:a}
indicated by the arrow pointing down (see \word{after the sale is
  completed}).  Having found the subordinate clause, we proceed to
extract the main clause by traversing the tree upwards and identifying
the S node immediately dominating the subordinate clause node (see the
arrow pointing up in Figure~\ref{fig:tree-extract:a}, \word{employees
  will lose their jobs}).  In cases where the subordinate clause is
sentence initial, we first identify the SBAR-TMP node and extract the
subtree dominated by it, and then traverse the tree downwards in order
to extract the S-tree immediately dominating it.

For the experiments described here we focus solely on subordinate
clauses immediately dominated by~S, thus ignoring cases where nouns
are related to clauses via a temporal marker (e.g.,~\word{John left
  after lunch}).  Note that there can be more than one main clause
that qualify as attachment sites for a subordinate clause. In
Figure~\ref{fig:tree-extract:a} the subordinate clause \word{after the
  sale is completed} can be attached either to \word{said} or
\word{will loose}.  There can be similar structural ambiguities for
identifying the subordinate clause; for example see
(\ref{eg:amb-conj}), where the conjunction \word{and} should lie
within the scope of the subordinate \word{before}-clause (and indeed,
the parser disambiguates the structural ambiguity correctly for this
case):

\begin{examples}
\item \label{eg:amb-conj} [ Mr. Grambling made off with \$250,000 of the
  bank's money [~before Colonial caught on and denied him the remaining
  \$100,000. ] ]
\end{examples}

We are relying on the parser for providing relatively accurate
resolutions of structural ambiguities, but unavoidably this will
create some noise in the data.  To estimate the extent of this noise,
we manually inspected 30 randomly selected examples for each of our
temporal discourse markers i.e.,~240 examples in total.  All the
examples that we inspected were true positives of temporal discourse
markers save one, where the parser assumed that \word{as} took a
sentential complement whereas in reality it had an NP complement
(i.e.,~\word{an anti-poverty worker}):

\begin{examples}
\item   \label{eg:false-pos}
[ He first moved to West Virginia [~as an anti-poverty worker, then
decided to stay and start a political career, eventually serving two
terms as governor. ] ]
\end{examples}

In most cases the noise is due to the fact that the parser either
overestimates or underestimates the extent of the text span for the
two clauses. 98.3\% of the main clauses and 99.6\% of the subordinate
clauses were accurately identified in our data set. Sentence
(\ref{eg:span-error}) is an example where the parser incorrectly
identifies the main clause: it predicts that the \word{after}-clause
is attached to \word{to denationalise the country's water industry}.
Note, however, that the subordinate clause (as \word{some managers
  resisted the move and workers threatened lawsuits}) is correctly
identified.

\begin{table}[t]
\begin{center}
\begin{tabular}{|l|r|d|}
\hline
TMark & Frequency & \multicolumn{1}{c |}{\mbox{ Distribution
    (\%)}} \\ \hline \hline 
  when    & 35,895        & 42.83        \\
  as      & 15,904       & 19.00        \\
  after   & 13,228       & 15.79        \\
  before  & 6,572        & 7.84       \\
  until   & 5,307        & 6.33       \\ 
  while   & 3,524         & 4.20       \\
  since   & 2,742        & 3.27       \\
  once    & 638          & 0.76       \\
\hline
TOTAL & 83,810 & 100.00 \\ \hline
\end{tabular}
\end{center}
\caption{Subordinate clauses extracted from {\sc Bllip} corpus}
\label{table:training-numbers}
\end{table}

\begin{examples}
\item \label{eg:span-error} [ Last July, the government postponed plans
  [ to denationalise the country's water industry [~after some managers
  resisted the move and workers threatened lawsuits. ] ] ]
\end{examples}
%
%
%
%
The size of the corpus we obtain with these extraction methods is
detailed in Table~\ref{table:training-numbers}.  There are 83,810
instances overall (i.e.,~just 0.20\% of the size of the corpus used by
Marcu and Echihabi, 2002). Also note that the distribution of temporal
markers ranges from 0.76\% (for \word{once}) to 42.83\% (for
\word{when}). 

Some discourse markers from our confusion set underspecify temporal
semantic information. For example, \word{when} can entail temporal
overlap (see (\ref{ex:when-amb}a), from \citeR{kamp:reyle:1993}), or
temporal progression (see (\ref{ex:when-amb}c), from
\citeR{moens:steedman:1988}). The same is true for \word{once},
\word{since}, and \word{as}:

\begin{examples}
\item   \label{ex:when-amb}
\begin{subexamples}
\item   Mary left when Bill was preparing dinner.\hfill ({\em temporal
    overlap}) 
\item   When they built the bridge, they solved all their traffic
  problems. \hfill ({\em temporal progression})
\end{subexamples}
\end{examples}

\begin{examples}
\item   \label{ex:once-amb}
\begin{subexamples}
\item   Once John moved to London, he got a job with the
  council.\hfill ({\em temporal progression})
\item   Once John was living in London, he got a job with
  the council. \hfill ({\em temporal overlap})
\end{subexamples}
\item   \label{ex:since-amb}
\begin{subexamples}
\item   John has worked for the council since he's been living in
  London. \hfill ({\em temporal overlap})
\item   John moved to London since he got a job with the council
  there.  \hfill ({\em cause} and hence {\em temporal precedence})
\end{subexamples}
\item \label{ex:as}
\begin{subexamples}
\item Grand melodies poured out of him as he contemplated Caesar's
  conquest of Egypt. \hfill ({\em temporal overlap})
\item I went to the bank as I ran out of cash. \hfill ({\em cause}, and
  hence {\em temporal precedence})
\end{subexamples}
\end{examples}

This means that if the model chooses \word{when}, \word{once}, or
\word{since} as the most likely marker between a main and subordinate
clause, then the temporal relation between the events described is
left underspecified.  Of course the semantics of \word{when} or
\word{once} limits the range of possible relations, but our model does
not identify which specific relation is conveyed by these markers for
a given example.  Similarly, \word{while} is ambiguous between a
temporal use in which it signals that the eventualities temporally
overlap (see (\ref{ex:while}a)) and a contrastive use which does not
convey any particular temporal relation (although such relations may
be conveyed by other features in the sentence, such as tense, aspect
and world knowledge; see (\ref{ex:while}b)).
\begin{examples}
\item \label{ex:while}
\begin{subexamples}
\item While the stock market was rising steadily, even companies
  stuffed with cash rushed to issue equity.
  
\item While on the point of history he was directly opposed to Liberal
  Theology, his appeal to a `spirit' somehow detachable from the
  Jesus of history run very much along similar lines to the Liberal
  approach.
\end{subexamples}
\end{examples}


We inspected 30 randomly-selected examples for markers with
underspecified readings (i.e.,~\word{when}, \word{once}, \word{since},
\word{while} and \word{as}). The marker \word{when} entails a temporal
overlap interpretation 70\% of the time and \word{as} entails temporal
overlap 75\% of the time, whereas \word{once} and \word{since} are
more likely to entail temporal progression (74\% and 80\%,
respectively). The markers \word{while} and \word{as} receive
predominantly temporal interpretations in our corpus.  Specifically,
\word{while} has non-temporal uses in~13.3\% of the instances in our
sample and \word{as} in~25\%. Once our interpretation model has been
applied, we could use these biases to disambiguate, albeit coarsely,
markers with underspecified meanings. Indeed, we demonstrate with
Experiment~\ref{exp:timebank} (see
Section~\ref{sec:timebank-experiment}) that our model is useful for
estimating {\em unambiguous} temporal relations, even when the
original sentence had no temporal marker, ambiguous or otherwise.




\subsection{Model Features}
\label{sec:feat-gener-temp}


A number of knowledge sources are involved in inferring temporal
ordering including tense, aspect, temporal adverbials, lexical
semantic information, and world knowledge
\shortcite{Asher:Lascarides:03}.  By selecting features that represent
these knowledge sources, notwithstanding indirectly and imperfectly,
we aim to empirically assess their contribution to the temporal
inference task. Below we introduce our features and provide motivation
behind their selection.

\paragraph{Temporal Signature (T)} It is well known that verbal tense
and aspect impose constraints on the temporal order of events and also
on the choice of temporal markers.  These constraints are perhaps best
illustrated in the system of \citeA{dorr:gaasterland:1995} who
examine how \emph{inherent} (i.e.,~states and events) and
\emph{non-inherent} (i.e.,~progressive, perfective) aspectual features
interact with the time stamps of the eventualities in order to
generate clauses and the markers that relate them.


Although we cannot infer inherent aspectual features from verb surface
form (for this we would need a dictionary of verbs and their aspectual
classes together with a process that assigns aspectual classes in a
given context), we can extract non-inherent features from our parse
trees. We first identify verb complexes including modals and
auxiliaries and then classify tensed and non-tensed expressions along
the following dimensions: finiteness, non-finiteness, modality,
aspect, voice, and polarity. The values of these features are shown in
Table~\ref{tab:tense}. The features finiteness and non-finiteness are
mutually exclusive.

\begin{table}[t]
\begin{center}
\begin{tabular}{|lll|} \hline
{\sc finite} & = & \{past, present\}\\
{\sc non-finite} &  = &\{0, infinitive, ing-form, en-form\}
\\  
{\sc modality} &  = &\{$\emptyset$, future, ability, possibility, obligation\} \\
{\sc aspect} &  = &\{imperfective, perfective, progressive\} \\
{\sc voice} &  = &\{active, passive\}\\
{\sc negation} &  = &\{affirmative, negative\}\\ \hline
\end{tabular}
\end{center}
\caption{\label{tab:tense} Temporal signatures} 
\end{table}

\begin{table}[t]
\begin{center}
\begin{tabular}{|lllll|} \hline
Feature       & once$_{\mathrm{M}}$  & once$_{\mathrm{S}}$ & since$_{\mathrm{M}}$
        & \word{since$_{\mathrm{S}}$}\\ \hline \hline
{\sc fin}     & 0.69  &  0.72 &  0.75 & 0.79\\
{\sc past}    & 0.28  &  0.34 &  0.35 & 0.71\\
{\sc act}     & 0.87  &  0.51 &  0.85 & 0.81\\
{\sc mod}     & 0.22  &  0.02 &  0.07 & 0.05\\
{\sc neg}  & 0.97  &  0.98 &  0.95 & 0.97\\\hline
\end{tabular}
\end{center}
\caption{\label{tab:signature} Relative frequency counts for temporal
        features in main (subscript~$\mathrm{M}$) and subordinate
        (subscript~$\mathrm{S}$) clauses}
\end{table}

Verbal complexes were identified from the parse trees heuristically by
devising a set of 30~patterns that search for sequences of
auxiliaries and verbs. From the parser output verbs were classified as
passive or active by building a set of 10~passive identifying patterns
requiring both a passive auxiliary (some form of \word{be} and
\word{get}) and a past participle.


To illustrate with an example, consider again the parse tree in
Figure~\ref{fig:tree-extract:a}. We identify the verbal groups
\word{will lose} and \word{is completed} from the main and
subordinate clause respectively. The former is mapped to the features
\{present, 0, future, imperfective, active, affirmative\}, whereas the
latter is mapped to \{present, 0, $\emptyset$, imperfective, passive,
affirmative\}, where~0 indicates the verb form is finite and
$\emptyset$ indicates the absence of a modal. In 
Table~\ref{tab:signature} we show the relative frequencies in our
corpus for finiteness ({\sc fin}), past tense ({\sc past}), active
voice ({\sc act}), and negation ({\sc neg}) for main and subordinate
clauses conjoined with the markers \word{once} and \word{since}.
As can be seen there are differences in the distribution of counts
between main and subordinate clauses for the same and different
markers. For instance, the past tense is more frequent in
\word{since} than \word{once} subordinate clauses and modal verbs are
more often attested in \word{since} main clauses when compared with
\word{once} main clauses. Also, \word{once} main clauses are more
likely to be active, whereas \word{once} subordinate clauses can be
either active or passive.

\paragraph{Verb Identity (V)}
Investigations into the interpretation of narrative discourse have
shown that specific lexical information plays an important role in
determining temporal interpretation
(e.g.,~\citeR{Asher:Lascarides:03}).  For example, the fact that verbs
like \word{push} can cause movement of their object and verbs like
\word{fall} describe the movement of their subject can be used to
interpret the discourse in~(\ref{ex:discourse}) as the pushing causing
the falling, thus making the linear order of the events mismatch their
temporal order.

\ex{
\item \label{ex:discourse}
Max fell. John pushed him. 
}


We operationalize lexical relationships among verbs in our data by
counting their occurrence in main and subordinate clauses from a
lemmatized version of the {\sc Bllip} corpus.  Verbs were extracted
from the parse trees containing main and subordinate clauses. Consider
again the tree in Figure~\ref{fig:tree-extract:a}.  Here, we identify
\word{lose} and \word{complete}, without preserving information about
tense or passivisation which is explicitly represented in our temporal
signatures.  Table~\ref{tab:verbs} lists the most frequent verbs
attested in main (Verb$_{\mathrm{M}}$) and subordinate
(Verb$_{\mathrm{S}}$) clauses conjoined with the temporal markers
\word{after, as, before, once, since, until, when}, and \word{while}
(TMark).

\begin{table}[t]
\begin{center}
\begin{tabular}{|l|ll|ll|ll|} \hline
TMark    & Verb$_{\mathrm{M}}$ & Verb$_{\mathrm{S}}$ &
Supersense$_{\mathrm{M}}$ &  Supersense$_{\mathrm{S}}$ &
Levin$_{\mathrm{M}}$ & Levin$_{\mathrm{S}}$\\\hline\hline
after   & sell   & leave    &  communication & communication & say & say \\   
as      & come   & acquire  & motion & motion & say & begin              \\              
before  & say    & announce & stative & stative & say & begin            \\           
once    & become & complete & stative & stative & say & get              \\             
since   & rise   & expect   & stative & change & say & begin             \\              
until   & protect& pay      & communication & possession & say & get     \\      
when    & make   & sell     & stative & motion &characterize  & get      \\        
while   & wait   & complete & communication & social & say & amuse       \\    \hline     
\end{tabular}
\end{center}
\caption{\label{tab:verbs} Most frequent verbs and verb classes in  main (subscript~$\mathrm{M}$) and subordinate  clauses (subscript~$\mathrm{M}$)} 
\end{table}

\paragraph{Verb Class (V$_{\mathrm{W}}$, V$_{\mathrm{L}}$)} The verb
identity feature does not capture meaning regularities concerning the
types of verbs entering in temporal relations. For example, in
Table~\ref{tab:verbs} \word{sell} and \word{pay} are possession verbs,
\word{say} and \word{announce} are communication verbs, and
\word{come} and \word{rise} are motion verbs.
\citeA{Asher:Lascarides:03} argue that many of the rules for
inferring temporal relations should be specified in terms of the {\em
  semantic class} of the verbs, as opposed to the verb forms
themselves, so as to maximize the linguistic generalizations captured
by a model of temporal relations. For our purposes, there is an
additional empirical motivation for utilizing verb classes as well as
the verbs themselves: it reduces the risk of sparse data.
Accordingly, we use two well-known semantic classifications for
obtaining some degree of generalization over the extracted verb
occurrences, namely WordNet \shortcite{fellbaum:1998} and the verb
classification proposed by \citeA{levin:1995}.


Verbs in WordNet are classified in~15 broad semantic domains
(e.g.,~verbs of change, verbs of cognition, etc.) often referred to as
supersenses \shortcite{Ciaramita:Johnson:03}. We therefore mapped the
verbs occurring in main and subordinate clauses to WordNet
supersenses (feature~V$_{\mathrm{W}}$). Semantically ambiguous verbs
will correspond to more than one semantic class. We resolve ambiguity
heuristically by always defaulting to the verb's prime sense (as
indicated in WordNet) and selecting its corresponding supersense. In
cases where a verb is not listed in WordNet we default to its
lemmatized form.


\citeA{levin:1995} focuses on the relation between verbs and
their arguments and hypothesizes that verbs which behave similarly with
respect to the expression and interpretation of their arguments share
certain meaning components and can therefore be organized into
semantically coherent classes (200 in total).
\citeA{Asher:Lascarides:03} argue that these classes provide
important information for identifying semantic relationships between
clauses.  Verbs in our data were mapped into their corresponding Levin
classes (feature~V$_{\mathrm{L}}$); polysemous verbs were
disambiguated by the method proposed by
\citeA{Lapata:Brew:04}.\footnote{\citeA{Lapata:Brew:04}
  develop a simple probabilistic model which determines for a given
  polysemous verb and its frame its most likely meaning overall
  (i.e.,~across a corpus), without relying on the availability of a
  disambiguated corpus.  Their model combines linguistic knowledge in
  the form of Levin (1995) classes and frame frequencies acquired from
  a parsed corpus.} Again, for verbs not included in Levin, the
lemmatized verb form was used. Examples of the most frequent Levin
classes in main and subordinate clauses as well as WordNet supersenses
are given in Table~\ref{tab:verbs}.


\paragraph{Noun Identity (N)}

It is not only verbs, but also nouns that can provide important
information about the semantic relation between two clauses;
\citeA{Asher:Lascarides:03} discuss an example in which having the
noun \word{meal} in one sentence and \word{salmon} in the other serves
to trigger inferences that the events are in a part-whole relation
(eating the salmon was part of the meal).  An example from our corpus
concerns the nouns \word{share} and \word{market}. The former is
typically found in main clauses preceding the latter which is often in
a subordinate clause. Table~\ref{tab:nouns} shows the most frequently
attested nouns (excluding proper names) in main (Noun$_{\mathrm{M}}$)
and subordinate (Noun$_{\mathrm{S}}$) clauses for each temporal
marker.  Notice that time denoting nouns (e.g.,~\word{year},
\word{month}) are relatively frequent in this data set.

\begin{table}[t]
\begin{center}
\begin{tabular}{|l|ll|ll|ll|} \hline
TMark   & Noun$_{\mathrm{N}}$ &  Noun$_{\mathrm{S}}$ & Supersense$_{\mathrm{M}}$ &  Supersense$_{\mathrm{S}}$  & Adj$_{\mathrm{M}}$ & Adj$_{\mathrm{S}}$\\\hline\hline
after   & year     &  company& act & act & last   & new \\                
as       & market   & dollar & act & act   & recent & previous       \\
before  & time     & year & act & group   & long   & new       \\
once    & stock    & place & act & act   & more   & new       \\
since   & company  & month & act & act    & first  & last     \\
until   & president& year & act & act    & new    & next   \\
when    & act & act& year     & year    & last   & last       \\
while   & group & act & chairman & plan    & first  & other
\\\hline   
\end{tabular}
\caption{\label{tab:nouns} Most frequent nouns, noun classes, and
  adjectives in  main (subscript~$\mathrm{M}$) and subordinate  clauses (subscript~$\mathrm{M}$)} 
\end{center}
\end{table}

Nouns were extracted from a lemmatized version of the {\sc Bllip}
corpus.  In Figure~\ref{fig:tree-extract:a} the nouns
\word{employees}, \word{jobs} and \word{sales} are relevant for the
Noun feature. In cases of noun compounds, only the compound head
(i.e.,~rightmost noun) was taken into account. A small set of rules
was used to identify organizations (e.g.,~\word{United Laboratories
  Inc.}), person names (e.g.,~\word{Jose Y. Campos}), and locations
(e.g.,~\word{New England}) which were subsequently substituted by the
general categories {\tt person}, {\tt organization}, and {\tt
  location}.

\paragraph{Noun Class (N$_{\mathrm{W}}$)}

As with verbs, \citeA{Asher:Lascarides:03} argue in favor of
symbolic rules for inferring temporal relations that utilize the
semantic classes of nouns wherever possible, so as to maximize the
linguistic generalizations that are captured.  For example, they argue
that one can infer a causal relation in (\ref{ex:bruise}) on the basis
that the noun \word{bruise} has a cause via some act-on predicate with
some underspecified agent (other nouns in this class include {\em
  injury}, \word{sinking}, \word{construction}):
\begin{examples}
\item   \label{ex:bruise} John hit Susan.  Her bruise is enormous.
\end{examples}
Similarly, inferring that \word{salmon} is part of a meal in
(\ref{ex:salmon}) rests on the fact that the noun \word{salmon}, in
one sense at least, denotes an edible substance.
\begin{examples}
\item   \label{ex:salmon}       John ate a wonderful meal.  He
  devoured lots of salmon.
\end{examples}

As in the case of verbs, nouns were also represented by supersenses
from the WordNet taxonomy. Nouns in WordNet do not form a single
hierarchy; instead they are partitioned according to a set of semantic
primitives into 25~supersenses (e.g.,~nouns of cognition, events,
plants, substances, etc.), which are treated as the unique beginners
of separate hierarchies. The nouns extracted from the parser were
mapped to WordNet classes.  Ambiguity was handled in the same way as
for verbs. Examples of the most frequent noun classes attested in main
and subordinate clauses are illustrated in Table~\ref{tab:nouns}.

\paragraph{Adjective (A)} 

Our motivation for including adjectives in the feature set is twofold.
First, we hypothesize that temporal adjectives (e.g.,~\word{old},
\word{new}, \word{later}) will be frequent in subordinate clauses
introduced by temporal markers such as \word{before}, \word{after},
and \word{until} and therefore may provide clues for relations
signaled by these markers. Secondly, similarly to verbs and nouns,
adjectives carry important lexical information that can be used for
inferring the semantic relation that holds between two clauses. For
example, antonyms can often provide clues about the temporal sequence
of two events (see \word{incoming} and \word{outgoing}
in~(\ref{ex:antonyms})).



\ex{
\item \label{ex:antonyms} 
The incoming president delivered his inaugural speech. The outgoing
president resigned last week.
}
As with verbs and nouns, adjectives were extracted from the parser's
output. The most frequent adjectives in main (Adj$_{\mathrm{M}}$) and
subordinate (Adj$_{\mathrm{S}}$) clauses are given in
Table~\ref{tab:verbs}.

\paragraph{Syntactic Signature (S)} The syntactic differences in main
and subordinate clauses are captured by the syntactic signature
feature.  The feature can be viewed as a measure of tree complexity,
as it encodes for each main and subordinate clause the number of NPs,
VPs, PPs, ADJPs, and ADVPs it contains. The feature can be easily read
off from the parse tree. The syntactic signature for the main clause
in Figure~\ref{fig:tree-extract:a} is [NP:2 VP:2 ADJP:0 ADVP:0 PP:0]
and for the subordinate clause [NP:1 VP:1 ADJP:0 ADVP:0 PP:0]. The
most frequent syntactic signature for main clauses is [NP:2 VP:1 PP:0
ADJP:0 ADVP:0]; subordinate clauses typically contain an adverbial
phrase [NP:2 VP:1 ADJP:0 ADVP:1 PP:0].  One motivating case for using
this syntactic feature involves verbs describing propositional
attitudes (e.g.,~\word{said}, \word{believe}, \word{realize}).  Our set
of temporal discourse markers will have varying distributions as to
their relative semantic scope to these verbs.  For example, one would
expect \word{until} to take narrow semantic scope (i.e.,~the {\em
  until}-clause would typically attach to the verb in the sentential
complement to the propositional attitude verb, rather than to the
propositional attitude verb itself), while the situation might be
different for \word{once}.

\paragraph{Argument Signature (R)} This feature captures the
argument structure profile of main and subordinate clauses. It applies
only to verbs and encodes whether a verb has a direct or indirect
object, and whether it is modified by a preposition or an adverbial.
As the rules for inferring temporal relations in
\citeA{hobbs:etal:1993} and \citeA{Asher:Lascarides:03}
attest, the predicate argument structure of clauses is crucial to
making the correct temporal inferences in many cases.  To take a
simple example, observe that inferring the causal relation in
(\ref{ex:discourse}) crucially depends on the fact that the subject of
\word{fall} denotes the same person as the direct object of \word{push};
without 
this, a relation other than a causal one would be inferred.

As
with syntactic signature, this feature was read from the main and
subordinate clause parse-trees. The parsed version of the {\sc Bllip}
corpus contains information about subjects. NPs whose nearest ancestor
was a VP were identified as objects.  Modification relations were
recovered from the parse trees by finding all PPs and ADVPs
immediately dominated by a VP.  In Figure~\ref{fig:tree-extract:a} the
argument signature of the main clause is [SUBJ~OBJ] and for the
subordinate it is [OBJ].

\paragraph{Position (P)} This feature simply records the position of
the two clauses in the parse tree, i.e.,~whether the subordinate
clause precedes or follows the main clause. The majority of the main
clauses in our data are sentence initial (80.8\%). However, there are
differences among individual markers. For example, \word{once} clauses
are equally frequent in both positions. 30\% of the \word{when}
clauses are sentence initial whereas 90\% of the \word{after} clauses
are found in the second position.  These statistics clearly show that
the relative positions of the main vs.\ subordinate clauses are going
to be relatively informative for the the interpretation task.

In the following sections we describe our experiments with the models
introduced in Section~\ref{sec:model}. We first investigate their
performance on temporal interpretation in the context of a
pseudo-disambiguation task (Experiment~\ref{exp:ml}). We also describe
a study with humans (Experiment~\ref{exp:human}) which enables us to
examine in more depth the models' behavior and the difficulty of the
inference task. Finally, we evaluate the proposed approach in a more
realistic setting, using sentences that do not contain explicit
temporal markers (Experiment~\ref{exp:timebank}).


\experiment{\label{exp:ml}}
\section{Experiment~\ref{exp:ml}: Temporal Inference as Pseudo-disambiguation}
\label{sec:exper-1:-conf}

\paragraph{Method}
Our models were trained on main and subordinate clauses extracted from
the {\sc Bllip} corpus as detailed in
Section~\ref{sec:parameter-estimation}. In the testing phase, all
occurrences of the relevant temporal markers were removed and the
models were used to select the marker which was originally attested in
the corpus. This experimental setup is admittedly artificial, but
important in revealing the difficulty of the task at hand. A model
that performs deficiently at the pseudo-disambiguation task, has
little hope of inferring temporal relations in a more natural setting
where events are neither connected via temporal markers nor found in a
main-subordinate relationship.

Recall that we obtained~83,810 main-subordinate pairs. These were
randomly partitioned into training (80\%), development (10\%) and test
data (10\%). Eighty randomly selected pairs from the test data were
reserved for the human study reported in Experiment~\ref{exp:human}.
We performed parameter tuning on the development set; all our results
are reported on the unseen test set, unless otherwise stated. We
compare the performance of the conjunctive and disjunctive models,
thereby assessing the effect of feature (in)dependence on the temporal
interpretation task.  Furthermore, we compare the performance of the
two proposed models against a baseline disjunctive model that employs
a word-based feature space (see~(\ref{eq:5}) where $P(a_{\langle M,i
  \rangle}=w_{\langle M,i\rangle}|t_{j})$) and $P(a_{\langle S,i
  \rangle}=w_{\langle S,i\rangle}|t_{j})$). This model resembles
\shortcitegen{Marcu:Echihabi:02}'s model in that it does not make use
of the linguistically motivated features presented in the previous
section; all that is needed for estimating its parameters is a corpus
of main-subordinate clause pairs. 
We also report the performance of a majority baseline (i.e.,~always
select \word{when}, the most frequent marker in our data set).

In order to assess the impact of our feature classes (see
Section~\ref{sec:feat-gener-temp}) on the interpretation task, the
feature space was exhaustively evaluated on the development set.  We
have nine classes, which results in $\frac{9!}{(9-k)!}$ combinations
where $k$ is the arity of the combination (unary, binary, ternary,
etc.). We measured the accuracy of all class combinations (1,023 in
total) on the development set.  From these, we selected the best
performing ones for evaluating the models on the test set.

\paragraph{Results} Our results are shown in Table~\ref{tab:summary}. We
report both accuracy and F-score. A set of diacritics is used to
indicate significance (on accuracy) throughout this paper (see
Table~\ref{tab:diacritics}).  The best performing disjunctive model on
the test set (accuracy 62.6\%) was observed with the combination of
verbs (V) with syntactic signatures (S).  The combination of verbs
(V), verb classes (V$_L$, V$_W$), syntactic signatures (S) and clause
position (P) yielded the highest accuracy (60.3\%) for the conjunctive
model.  Both conjunctive and disjunctive models performed
significantly better than the majority baseline and word-based model
which also significantly outperformed the majority baseline. The
disjunctive model (SV) significantly outperformed the conjunctive one
(V$_{\mathrm{W}}$V$_{\mathrm{L}}$PSV).

\begin{table}
\begin{center}
\begin{tabular}{|c|l|}\hline
\multicolumn{1}{|c|}{Symbols} & \multicolumn{1}{c|}{Meaning}  \\ \hline
$*$     &  significantly different from Majority Baseline  \\
$\dag$  &  significantly different from Word-based Baseline  \\
\$      & significantly different from Conjunctive Model \\
$\ddag$ &  significantly different from Disjunctive Model  \\
$\&$    & significantly different from Conjunctive Ensemble\\ 
$\#$    &  significantly different from Disjunctive
                  Ensemble\\ \hline
\end{tabular}
\end{center}
\caption{\label{tab:diacritics} Meaning of diacritics indicating
  statistical significance ($\chi^2$ tests, \mbox{$p < 0.05$})}
\end{table}

\begin{table}
\begin{center}
\begin{tabular}{|l|f|c|} \hline
Model                      &     \labe{Accuracy} &  \labe{F-score} \\\hline
Majority Baseline          &    42.6^{\dag\$\ddag\#\&} &   NA    \\
Word-based Baseline        &    48.2^{*\$\ddag\#\&}    &   44.7  \\
Conjunctive (V$_{\mathrm{W}}$V$_{\mathrm{L}}$PSV)        & 60.3^{*\dag\ddag\#\&} &   53.3  \\
Disjunctive (SV)           &    62.6^{*\dag\$\#\&} &   62.3  \\
\hline \hline              
Ensemble (Conjunctive)     &    64.5^{*\dag\$\ddag\&} &   59.9  \\ 
Ensemble (Disjunctive)     &    70.6^{*\dag\$\ddag\#} &   69.1   \\  \hline
\end{tabular}
\end{center}
\caption{\label{tab:summary} Summary of results for the temporal
  pseudo-disambiguation task; comparison of
  baseline models against conjunctive and disjunctive models and their
  ensembles (V: verbs, V$_{\mathrm{W}}$:  WordNet verb supersenses,
  V$_{\mathrm{L}}$: Levin verb classes, P: clause position, S:
  syntactic signature)} 
\end{table}

We attribute the conjunctive model's worse performance to data
sparseness.  There is clearly a trade-off between reflecting the true
complexity of the task of inferring temporal relations and the amount
of training data available.  The size of our data set favors a
simpler model over a more complex one.  The difference in performance
between the models relying on linguistically-motivated features and
the word-based model also shows that linguistic abstractions are
useful in overcoming sparse data.

 
 
\begin{figure}[t!]
  \centering 
  \includegraphics[width=9cm]{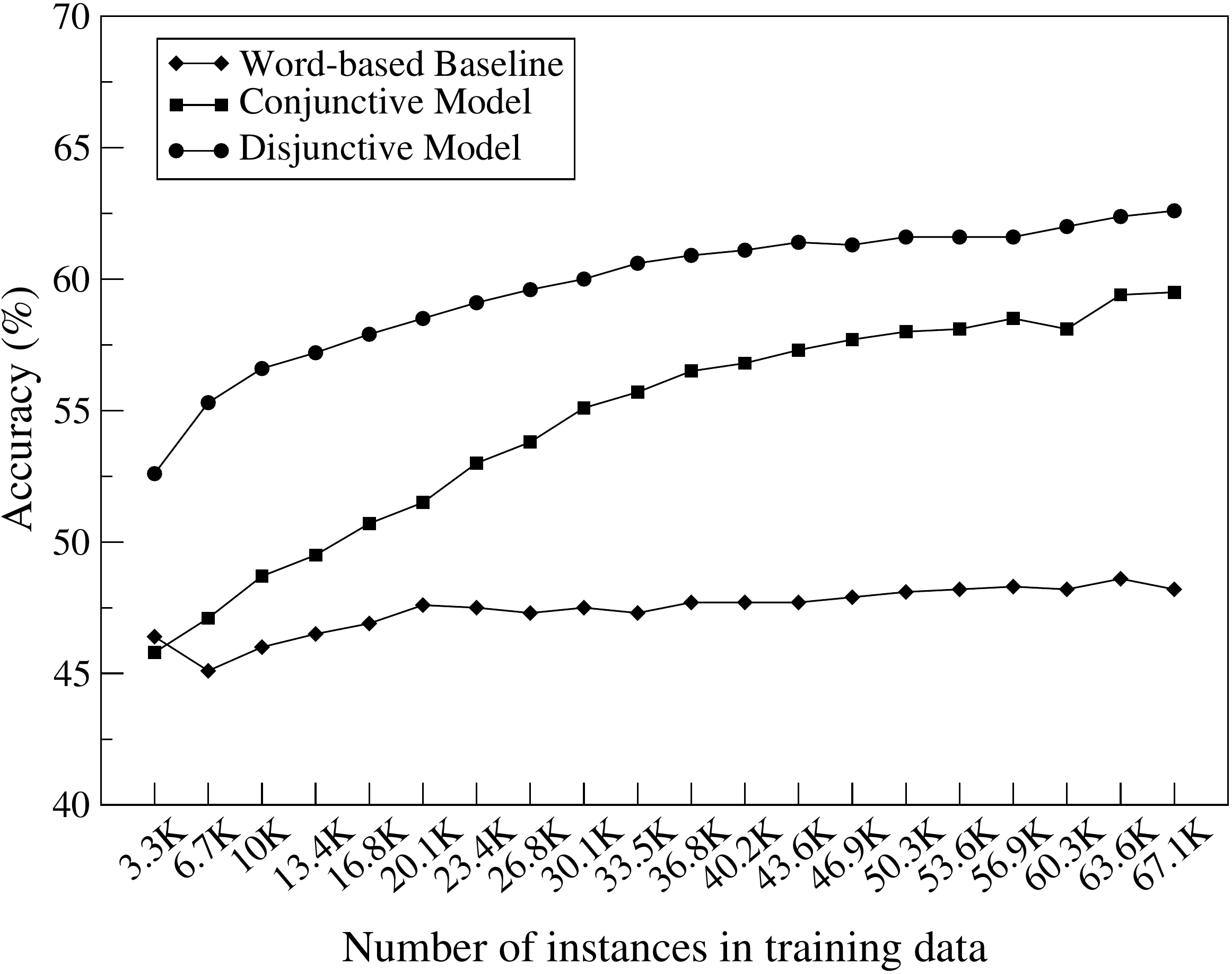}
  \caption{Learning curve for conjunctive, disjunctive, and word-based
  models.
   \label{fig:learning}}
 \end{figure}

We further analyzed the data requirements for our models by varying
the amount of instances on which they are trained.
Figure~\ref{fig:learning} shows learning curves for the best
conjunctive and disjunctive models
(V$_{\mathrm{W}}$V$_{\mathrm{L}}$PSV and SV). For comparison, we also
examine how training data size affects the (disjunctive) word-based
baseline model.  As can be seen, the disjunctive model has an
advantage over the conjunctive one; the difference is more pronounced
with smaller amounts of training data. Very small performance gains
are obtained with increased training data for the word baseline model.
A considerably larger training set is required for this model to be
competitive against the more linguistically aware models. This result
is in agreement with \citeA{Marcu:Echihabi:02} who employ a very large
corpus (1 billion words, from which they extract 40 million training
examples) for training their word-based model.

 Further analysis of our models' output revealed that some feature
 combinations performed reasonably well on individual markers for both
 the disjunctive and conjunctive model, even though their overall
 accuracy did not match the best feature combinations for either model
 class.  Some accuracies for these combinations are shown in
 Table~\ref{tab:dev:features}. For example, NPRSTV was one of the best
 combinations for generating \word{after} under the disjunctive model,
 whereas SV was better for \word{before} (feature abbreviations are as
 introduced in Section~\ref{sec:feat-gener-temp}). Given the
 complementarity of different models, an obvious question is whether
 these can be combined. An important finding in machine learning is
 that a set of classifiers whose individual decisions are combined in
 some way (an \word{ensemble}) can be more accurate than any of its
 component classifiers if the errors of the individual classifiers are
 sufficiently uncorrelated \shortcite{Dietterich:97}. The next section
 reports on our ensemble learning experiments.

\begin{table}
\begin{center}
\begin{tabular}{|l|ld|ld|} \hline
      & \multicolumn{2}{c}{Disjunctive Model}  &
      \multicolumn{2}{c|}{Conjunctive Model}\\
TMark   & Features   &  \labe{Accuracy}    & Features   & \labe{Accuracy}  \\ \hline \hline
after & NPRSTV & 69.9     & V$_{\mathrm{W}}$PTV   & 79.6 \\
as    & ANN$_{\mathrm{W}}$PSV & 57.0 & V$_{\mathrm{W}}$V$_{\mathrm{L}}$SV & 57.0 \\
before& SV     & 42.1   & TV & 11.3 \\
once  &  PRS   & 40.7   & V$_{\mathrm{W}}$P & 3.7 \\
since &  PRST  & 25.1   & V$_{\mathrm{L}}$V & 1.0 \\
when  & V$_{\mathrm{L}}$PS & 85.5 & V$_{\mathrm{L}}$NV & 86.5\\ 
while & PST    & 49.0   & V$_{\mathrm{L}}$PV & 9.6 \\
until & V$_{\mathrm{L}}$V$_{\mathrm{W}}$RT & 69.4 & V$_{\mathrm{W}}$V$_{\mathrm{L}}$PV & 9.5\\ \hline
\end{tabular}
\end{center}
\caption{\label{tab:dev:features} Best feature combinations for
  individual markers (development set; V:
  verbs, V$_{\mathrm{W}}$: WordNet verb supersenses,
  V$_{\mathrm{L}}$: Levin verb classes, N: nouns, 
  N$_{\mathrm{W}}$: WordNet noun supersenses, P:
  clause position, R: argument signature, S: syntactic signature, T:
  tense signature)}
\end{table}

\paragraph{Ensemble Learning}

An ensemble of classifiers is a set of classifiers whose individual
decisions are combined to classify new examples. This simple idea has
been applied to a variety of classification problems ranging from
optical character recognition to medical diagnosis and part-of-speech
tagging (for overviews, \citeR<see>{Dietterich:97,Halteren:ea:01}).
Ensemble learners often yield superior results to individual learners
provided that the component learners are accurate and diverse
\shortcite{Hansen:Salamon:90}.

An ensemble is typically built in two steps: first multiple component
learners are trained and next their predictions are combined.
Multiple classifiers can be generated either by using subsamples of
the training data \shortcite{Breiman:96a,Freund:Shapire:96} or by
manipulating the set of input features available to the component
learners \shortcite{Cherkauer:96}.  Weighted or unweighted voting is
the method of choice for combining individual classifiers in an
ensemble. A more sophisticated combination method is \emph{stacking}
where a learner is trained to predict the correct output class when
given as input the outputs of the ensemble classifiers
\shortcite{Wolpert:92,Breiman:96b,Halteren:ea:01}. In other words, a
second-level learner is trained to select its output on the basis of
the patterns of co-occurrence of the output of several component
learners.

We generated multiple classifiers (for combination in the ensemble) by
varying the number and type of features available to the conjunctive
and disjunctive models discussed in the previous section. The outputs
of these models were next combined using c5.0 \shortcite{Quinlan:93},
a decision-tree second level-learner. Decision trees are among the
most widely used machine learning algorithms. They perform a general
to specific search of a feature space, adding the most informative
features to a tree structure as the search proceeds. The objective is
to select a minimal set of features that efficiently partitions the
feature space into classes of observations and assemble them into a
tree (for details, see \citeR{Quinlan:93}). A classification for a
test case is made by traversing the tree until either a leaf node is
found or all further branches do not match the test case, and
returning the most frequent class at the last node.

Learning in this framework requires a primary training set for
training the component learners; a secondary training set for training
the second-level learner and a test set for assessing the stacked
classifier. We trained the decision-tree learner on the development
set using 10-fold cross-validation. We experimented with 133~different
conjunctive models and~65 disjunctive models; the best results on the
development set were obtained with the combination of 22~conjunctive
models and 12~disjunctive models. The component models are presented
in Table~\ref{tab:compmodels}.  The ensembles' performance on the test
set is reported in Table~\ref{tab:summary}.

\begin{table}
\begin{center}
\begin{tabular}{|l@{\hspace{0.4em}}l@{\hspace{0.4em}}l@{\hspace{0.4em}}l@{\hspace{0.4em}}l@{\hspace{0.4em}}l@{\hspace{0.4em}}l|} \hline
\multicolumn{7}{|c|}{Conjunctive Ensemble} \\\hline
APTV &  PSVV$_{\mathrm{W}}$N$_{\mathrm{W}}$V$_{\mathrm{L}}$ & NPVV$_{\mathrm{W}}$V$_{\mathrm{L}}$ & PRTVV$_{\mathrm{W}}$V$_{\mathrm{L}}$ &PSV$_{\mathrm{W}}$V$_{\mathrm{L}}$ & PSVV$_{\mathrm{W}}$V$_{\mathrm{L}}$ & PVV$_{\mathrm{W}}$V$_{\mathrm{L}}$ \\
SVV$_{\mathrm{W}}$V$_{\mathrm{L}}$&  NSVV$_{\mathrm{W}}$ &  PSVV$_{\mathrm{W}}$ 
PVV$_{\mathrm{W}}$ &  N$_{\mathrm{W}}$PSVV$_{\mathrm{L}}$ & PSV$_{\mathrm{L}}$ &
 PVV$_{\mathrm{L}}$ &  NPSV  \\
  NPV & NSV & PSV &  PV &  SV & TV &  V \\\hline
\multicolumn{7}{|c|}{Disjunctive Ensemble} \\ \hline
AN$_{\mathrm{W}}$NPSV & APSV &  ASV &  PRSV$_{\mathrm{W}}$ &
PSV$_{\mathrm{N}}$& SV$_{\mathrm{L}}$ & NPRSTV \\
 PRS &  PRST & PRSV& PSV &  SV & &\\  \hline
\end{tabular}
\caption{\label{tab:compmodels} Component models for ensemble
  learning (A: adjectives, V: verbs, V$_{\mathrm{W}}$: WordNet verb
  supersenses, V$_{\mathrm{L}}$: Levin verb classes, N: nouns,
  N$_{\mathrm{W}}$: WordNet noun supersenses, P: clause position, R:
  argument signature, S: syntactic signature, T: tense signature)}
\end{center}
\end{table}

As can be seen, both types of ensemble significantly outperform the
word-based baseline, and the best performing individual models.
Furthermore, the disjunctive ensemble significantly outperforms the
conjunctive one.  Table~\ref{tab:ensemble:details} details the
performance of the two ensembles for each individual marker. 
Both ensembles have difficulty inferring the markers \word{since},
\word{once} and \word{while}; the difficulty is more pronounced in the
conjunctive ensemble.  We believe that the worse performance for
predicting these relations is due to a combination of sparse data and
ambiguity.  First, observe that these three classes have fewest
examples in our data set (see Table~\ref{table:training-numbers}).
Secondly, \word{once} is temporally ambiguous, conveying temporal
progression and temporal overlap (see example~(\ref{ex:once-amb})).
The same ambiguity is observed with \word{since} (see
example~(\ref{ex:since-amb})).  Finally, although the temporal sense
of {\em while} always conveys temporal overlap, it has a non-temporal,
contrastive sense too which potentially creates some noise in the
training data, as discussed in Section~\ref{sec:data-extraction}.
Another contributing factor to \word{while}'s poor performance is the
lack of sufficient training data. Note that the extracted instances
for this marker constitute only 4.2\% of our data.  In fact, the model
often confuses the marker \word{since} with the semantically similar
\word{while}. This could be explained by the fact that the majority of
training examples for {\em since} had interpretations that imply
temporal overlap, thereby matching the temporal relation implied by
{\em while}, which in turn was also the majority interpretation in our
training corpus (the non-temporal, contrastive sense accounting for
only 13.3\% of our training examples).


\begin{table}
\begin{center}
\begin{tabular}{|l|d|d|d|d|}\hline 
      & \multicolumn{2}{c|}{Disjunctive Ensemble}  &
      \multicolumn{2}{c|}{Conjunctive Ensemble}  \\
TMark &  \labe{Accuracy}  &  \labe{F-score}   & \labe{Accuracy}  & \labe{F-score}    \\\hline\hline  
after &  66.4 &    63.9    & 59.3 & 57.6  \\  
as    &  62.5 &    62.0    & 59.0 & 55.1  \\  
before&  51.4 &    50.6    & 17.1 & 22.3  \\  
once  &  24.6 &    35.3    & 0.0  & 0.0  \\  
since &  26.2 &    38.2    & 3.9  & 4.5  \\  
when  &  91.0 &    86.9    & 90.5 & 84.7  \\  
while &  28.8 &    41.2    & 11.5 & 15.8  \\  
until &  47.8 &    52.4    & 17.3 & 24.4  \\  \hline \hline
All   &  70.6 &    69.1    & 64.5 & 59.9  \\   \hline
\end{tabular}
\end{center}
\caption{\label{tab:ensemble:details} Ensemble results on sentence
      interpretation for individual markers (test set)}
\end{table}

Let us now examine which classes of features have the most impact on
the interpretation task by observing the component learners selected
for our ensembles. As shown in Table~\ref{tab:dev:features}, verbs
either as lexical forms (V) or classes (V$_{\mathrm{W}}$,
V$_{\mathrm{L}}$), the syntactic structure of the main and subordinate
clauses (S) and their position (P) are the most important features for
interpretation. Verb-based features are present in all component
learners making up the conjunctive ensemble and in 10 (out of 12)
learners for the disjunctive ensemble. The argument structure feature
(R) seems to have some influence (it is present in five of the~12
component (disjunctive) models), however we suspect that there is some
overlap with~S.  Nouns, adjectives and temporal signatures seem to
have a small impact on the interpretation task, at least in the WSJ
domain.  Our results so far point to the importance of the lexicon for
inferring temporal relations but also indicate that the syntactic
complexity of the two clauses is an another key predictor.  Asher and
Lascarides' \citeyear{Asher:Lascarides:03} symbolic theory of
discourse interpretation also emphasizes the importance of lexical
information in inferring temporal relations, while Soricut and Marcu
\citeyear{Soricut:Marcu:03} find that syntax trees are useful for
inferring discourse relations, some of which have temporal
consequences.

\experiment{\label{exp:human}}
\section{Experiment~\ref{exp:human}: Human Evaluation}
\label{sec:experiment-2:-human}

\paragraph{Method}

We further assessed the temporal interpretation model by comparing its
performance against human judges.  Participants were asked to perform
a multiple choice task.  They were given a set of~40 main-subordinate
pairs (five for each marker) randomly chosen from our test data. The
marker linking the two clauses was removed and participants were asked
to select the missing word from a set of eight temporal markers, thus
mimicking the model's task. Examples of the materials our participants
saw are given in Apendix~A.
 
The study was conducted remotely over the Internet.  Subjects first
read a set of instructions that explained the task, and had to fill in
a short questionnaire including basic demographic information.  A
random order of main-subordinate pairs and a random order of markers
per pair was generated for each subject. The study was completed
by~198 volunteers, all native speakers of English.  Subjects were
recruited via postings to local Email lists.

\paragraph{Results}

Our results are summarized in Table~\ref{tab:agreement}. We measured
how well our subjects (Human) agree with the gold standard
(Gold)---i.e.,~the corpus from which the experimental items were
selected---and how well they agree with each other (Human-Human). We
also show how well the disjunctive ensemble (Ensemble) agrees with the
subjects (Ensemble-Human) and the gold standard (Ensemble-Gold).  We
measured agreement using the Kappa coefficient
\shortcite{Siegel:Castellan:88} but also report percentage agreement
to facilitate comparison with our model.  In all cases we compute
pairwise agreements and report the mean.

\begin{table}[t]
\begin{center}
\begin{tabular}{|l|c|c|c|c|} \hline
      & $K$        & \%                   \\ \hline\hline  
Human-Human      & .410       & 45.0                  \\ 
Human-Gold       & .421       & 46.9                  \\ 
Ensemble-Human   & .390       & 44.3                  \\ 
Ensemble-Gold    & .413       & 47.5                  \\ \hline 
\end{tabular}
\end{center}
\caption{\label{tab:agreement} Agreement figures for subjects and
  disjunctive ensemble (Human-Human: inter-subject agreement, Human-Gold: agreement
  between subjects and gold standard corpus, Ensemble-Human: agreement between ensemble
  and subjects, Ensemble-Gold: agreement between ensemble and gold
  standard corpus)}
\end{table}

\begin{table}[t]
  \begin{center}
  \begin{tabular}{|l|cccccccc|} \hline
         & after &  as &  before &   once &  since &  until & when &  while\\ \hline
  after  & {\bf .55}   & .06 & .03     &  .10   & .04    & .01    & .20   & .01\\
  as     & .14   & {\bf .33} & .02     &  .02   & .03    & .03    & .20   &  .23\\
  before & .05   & .05 & {\bf .52} &  .08   & .03    & .15    & .08   & .04\\
  once   & .17   & .06 & .10     &  {\bf .35}   & .07    & .03    & .17   & .05\\
  since  & .10   & .09 & .04     &  .04   & {\bf .63}    & .03    & .06   & .01\\
  until  & .06   & .03 & .05     &  .10   & .03    & {\bf .65}    & .05   & .03\\
  when   & .20   & .07 & .09     &  .09   & .04    & .03    & {\bf .45}   & .03\\
  while  & .16   & .05 & .08     &  .03   & .04    & .02    & .10   & {\bf .52}\\\hline
  \end{tabular}
\caption{\label{tab:confmatrix} Confusion matrix based on percent
  agreement between subjects} 
  \end{center}
  \end{table}

  As shown in Table~\ref{tab:agreement} there is moderate
  agreement\footnote{\citeA{Landis:Koch:77} give the following five
    qualifications for different values of Kappa: .00--.20 is slight,
    .21--.40 is fair, .41--.60 is moderate, .61--.80 is substantial,
    whereas .81--1.00 is almost perfect.} among humans when selecting
  an appropriate temporal marker for a main and a subordinate clause.
  The ensemble's agreement with the gold standard approximates human
  performance on the interpretation task ($K = .413$ for~Ensemble-Gold
  vs.\ $K = .421$ for~Human-Gold).  The agreement of the ensemble with
  the subjects is also close to the upper bound, i.e.,~inter-subject
  agreement (see~Ensemble-Human and Human-Human in
  Table~\ref{tab:agreement}). Further analysis revealed that the
  majority of disagreements among our subjects arose for \word{as} and
  \word{once} clauses.  \word{Once} was also problematic for the
  ensemble model (see Table~\ref{tab:ensemble:details}).  The
  inter-subject agreement was~33\% for \word{as} clauses and 35\% for
  \word{once} clauses.  For the other markers, the subject agreement
  was around 55\%.  The highest agreement was observed for
  \word{since} and \word{until} (63\% and 65\% respectively). A
  confusion matrix summarizing the resulting inter-subject agreement
  for the interpretation task is shown in Table~\ref{tab:confmatrix}.

  The moderate agreement is not entirely unexpected given that some of
  the markers are semantically similar and in some cases more than one
  marker are compatible with the temporal implicatures that arise from
  joining the two clauses. For example, \word{when} can be compatible
  with \word{after}, \word{as}, \word{before}, \word{once}, and
  \word{since}. Besides \word{when}, \word{as} can be compatible with
  \word{since}, and \word{while}. Consider for example the following
  sentence from our experimental materials: \word{More and more older
    women are divorcing when their husbands retire}. Although
  \word{when} is the right connective according to the corpus,
  \word{once} or \word{after} are also valid choices. Indeed
  \word{after} is often chosen instead of \word{when} by our subjects
  (see Table~\ref{tab:confmatrix}).  Also note that neither the model
  nor the subjects have access to the context surrounding the sentence
  whose marker must be inferred. In the sentence \word{A lot of them
    want to get out before they get kicked out} (again taken from our
  materials), knowing the referents of \word{them} and \word{they} is
  important in selecting the right relation. In some cases,
  substantial background knowledge is required to make a valid
  temporal inference. In the sentence \word{Are more certified deaths
    required before the FDA acts?} (see Appendix~A), one must know
  what FDA stands for (i.e.,~Federal, Food, Drug, and Cosmetic Act).
  In a less strict evaluation setting where more than one connective
  are considered correct (on the basis of semantic compatibility), the
  inter-subject agreement is $K=.640$ (67.7\%). Moreover, the
  ensemble's agreement with the subjects is $K=.609$ (67\%).

  We next evaluate the performance of the ensemble model on a more
  challenging task. Our test data so far has been somewhat
  artificially created by removing the temporal marker connecting a
  main and subordinate clause. Although this experimental setup allows
  to develop and evaluate temporal inference models relatively
  straightforwardly, it remains unsatisfactory. In most cases a
  temporal model would be required for interpreting events that are
  not only attested in main-subordinate clauses but in a variety of
  constructions (e.g.,~in parataxis or indirect speech) which may not
  contain temporal markers. We use the annotations in the TimeBank
  corpus for investigating whether our model, which is trained on
  automatically annotated data, performs well on a more realistic test
  set.


\experiment{\label{exp:timebank}}
\section{Experiment~\ref{exp:timebank}: Predicting TimeML Relations}
\label{sec:timebank-experiment}

\paragraph{Method}

As mentioned earlier the TimeBank corpus has been manually annotated
with the TimeML coding scheme. In this scheme, verbs, adjectives, and
nominals are annotated as {\sc event}s and are marked up with
attributes such as the class of the event (e.g.,~state, reporting),
its tense (e.g.,~present, past), aspect (e.g., perfective,
progressive), and polarity (positive or negative). The {\sc tlink} tag
is used to represent temporal relationships between events, or between
an event and a time. These relationships can be inter- or
intra-sentential. Table~\ref{tab:tlinkrels} illustrates the {\sc
  tlink} relationships with sentences taken from the TimeBank corpus.
We focus solely on intra-sentential temporal relations between events;
Table~\ref{tab:tlinkrels} does not include the {\sc identity}
relationship which is commonly attested inter-sententially.


\begin{table}[t]
\begin{center}
\begin{tabular}{|l@{\hspace{0.1em}}lp{10cm}|} \hline
{\sc before} & ({\sc before})& Pacific First Financial Corp. \textbf{said} shareholders \textbf{approved}
its acquisition by Royal Trusstco Ltd. of Toronto for \$27 a share, or
\$212 million. \\ \hline
{\sc ibefore}&  ({\sc before})& The first would be to  launch the much-feared direct  invasion of
Saudi Arabia, hoping to \textbf{seize} some Saudi oil fields and
\textbf{improve} his bargaining position.  \\\hline
{\sc after} & ({\sc before})& In Washington today  the Federal Aviation Administration
\textbf{released} air traffic control tapes from the night the TWA flight eight
hundred \textbf{went} down. \\\hline
{\sc iafter} & ({\sc before})& In addition, Hewlett-Packard \textbf{acquired} a
two-year option to \textbf{buy} an extra 10\%, of which half may be
sold directly to Hewlett-Packard by Octel. \\\hline
{\sc includes} & (\sc includes)& Under the offer, shareholders  will \textbf{receive}
one right for each 105  common shares \textbf{owned}.  \\ \hline
{\sc is\_included}&  (\sc includes) & The purchase price was \textbf{disclosed} in a
preliminary prospectus \textbf{issued} in connection with MGM Grand's
planned offering of six million common shares. \\\hline
{\sc during} & ({\sc includes}) &  According to Jordanian officials, a smaller line into
Jordan  \textbf{remained} \textbf{operating}. \\ \hline
{\sc ends}&  ({\sc ends})& The government may  move to \textbf{seize} the money that
Mr. Antar is \textbf{using}  to pay legal fees.  \\\hline
{\sc ended\_by}&  ({\sc ends})& The Financial Times 100-share index \textbf{shed} 47.3 points to
\textbf{close} at 2082.1, down 4.5\% from the previous Friday. \\ \hline
{\sc begins}&  ({\sc begins})& DPC, an investor group led by New York-based Crescott Investment
Associates, had itself \textbf{filed} a suit in state court in Los
Angeles \textbf{seeking} to nullify the agreement. \\\hline
{\sc begun\_by} &  ({\sc begins})& Saddam said he will \textbf{begin} \textbf{withdrawing} troops from
Iranian territory on Friday and release Iranian prisoners of war. \\\hline
{\sc simultaneous} & & Nearly 200 Israeli soldiers have been \textbf{killed}
\textbf{fighting} Hezbollah and other guerrillas guerrillas.\\ \hline
\end{tabular} 
\end{center}
\caption{\label{tab:tlinkrels} {\sc tilink} relationships in TimeBank;
the events participating in the relationship are marked with boldface;
a more coarse-grained set of relationships is shown within
parentheses. }
\end{table}

Our intent here is to use the model presented in the previous sections
to interpret the temporal relationships between events like those
shown in Table~\ref{tab:tlinkrels} in the absence of overtly
verbalized temporal information (e.g.,~temporal markers).  However,
one stumbling block to performing this kind of evaluation is that the
corpus on which our model was trained uses different labels from those
in Table~\ref{tab:tlinkrels} (e.g.,~(ambiguous) temporal markers like
\word{when}).  Fortunately, the temporal markers we considered and the
TimeML relations are more or less semantically compatible, and so a
mapping can be devised. First notice that some of the relations in
Table~\ref{tab:tlinkrels} are redundant. For instance {\sc before} is
the inverse of {\sc after}, {\sc is\_included} is the inverse of {\sc
  includes}, and so on. Furthermore, some semantic distinctions are
too fine-grained for our model to identify accurately (e.g.,~{\sc
  before} and {\sc ibefore} (immediately before), {\sc simultaneous}
and {\sc during}). We therefore reduced the relations in
Table~\ref{tab:tlinkrels} into a smaller set by collapsing {\sc
  before}, {\sc ibefore}, {\sc after} and {\sc iafter} (immediately
after) into one relationship. Analogously, we collapsed {\sc
  simultaneously} and {\sc during}, {\sc includes} and {\sc
  is\_included}, {\sc begins} and {\sc begun\_by}, and {\sc ends} and
{\sc ended\_by}. The reduced relation set is also shown in
Table~\ref{tab:tlinkrels} (within parentheses).

\begin{table}[t]
\begin{center}
\begin{tabular}{|l|l|t|t|} \hline
TMark                          & TimeMLRel & \labe{TrainInst} & \labe{TestInst}\\\hline\hline 
after,before,once,when   &  {\sc before}   & 31,643    &  877\\
as,when,while     & {\sc includes}         & 21,859    & 246\\
as,when,while     & {\sc simultaneous}     & 22,165    & 360\\
since                   & {\sc begins}     & 2,810     & 19\\
until                   & {\sc ends}       & 5,333     & 64\\ 
no-temp-rel             & {\sc no-temp-rel} & 22,523   & 967\\ \hline
\end{tabular}
\caption{\label{tab:mapping} Mapping between temporal markers and
  coarse-grained set of TimeML relations; number of training and test
  instances per relation.}
\end{center}
\end{table}

We next defined a mapping between our temporal connectives and the
reduced set of TimeML relations (see Table~\ref{tab:mapping}). Such a
mapping cannot be one-to-one, since some of our connectives are
compatible with more than one temporal relationship (see
Section~\ref{sec:data-extraction}). For instance \word{when} can
indicate an {\sc includes} or {\sc before} relationship. We also
expect this mapping to be relatively noisy given that some temporal
markers entail non-temporal relationships (e.g.,~\word{while}).
Table~\ref{tab:mapping} includes an additional relation, namely
``no-temp-rel''. We thus have the option of not assigning any temporal
relation, thereby avoiding the pitfall of making a wrong prediction in
cases where non-temporal inferences are entailed by any two events. We
next describe how training and test instances were generated for our
experiments.

The disjunctive ensemble model from Experiment~\ref{exp:ml} was
trained on the {\sc Bllip} corpus using the same features and
component learners described in Sections~\ref{sec:feat-gener-temp}
and~\ref{sec:exper-1:-conf}. The training data consisted of our
original 83,810 main-subordinate clause pairs labeled with the
temporal relations from Table~\ref{tab:mapping} (second column). To
these we added 22,523 instances representative of the {\sc
  no-temp-rel} relation. Such instances were gathered by randomly
concatenating main and subordinate clauses belonging to different
documents (for a similar method, see \citeR{Marcu:Echihabi:02}). We
hypothesize that the two clauses do not trigger temporal relations,
since they are neither syntactically nor semantically related.
Instances with connectives \word{since} and \word{once} were mapped to
labels {\sc begins} and {\sc ends}, respectively. In addition to {\sc
  begins}, \word{since} can signal {\sc before}, {\sc includes}, and
{\sc simultaneous} temporal relations.  However, in our experiments
instances with \word{since} were used to exclusively learn the {\sc
  begins} relation.  This is far from perfect, but we felt necessary
since {\sc begins} is not represented by any other temporal marker.
The training instances for \word{as} and \word{while} were equally
split between the relationships {\sc includes} and {\sc simultaneous}.
Similarly, the data for \word{when} was equally split among {\sc
  before}, {\sc includes}, and {\sc simultaneous}. Instances with
\word{after}, \word{before}, and \word{once} were exclusively used for
learning the {\sc before} relation.  The number of training instances
per relation (TrainInst) is given in Table~\ref{tab:mapping}.

As test data, we used sentences from the TimeBank corpus. We only
tested the ensemble model on intra-sentential event-event relations.
Furthermore, we excluded sentences with overt temporal connectives, as
we did not want to positively influence the model's performance. The
TimeBank corpus is not explicitly annotated with the {\sc no-temp-rel}
relation. There are however sentences in the corpus whose events do
not participate in any temporal relationship. We therefore
hypothesized that these sentences were representative of {\sc
  no-temp-rel}. The total number of test instances (TestInst) used in
this experiment is given in
Table~\ref{tab:mapping}.

\begin{table}[t]
\begin{center}
\begin{tabular}{|l|c|d|c|} \hline
Model                  & TrainCorpus & \labe{Accuracy} & \labe{F-score} \\ \hline
Majority Baseline      & {\sc Bllip} & 34.7         & NA  \\
Word-based Baseline    & {\sc Bllip} & 39.1^{*}     & 21.1  \\
Ensemble (Disjunctive) & {\sc Bllip} & 53.0^{*\dag} & 45.8  \\
Ensemble (Disjunctive) & TimeBank    & 42.7         & 40.5 \\ \hline
\end{tabular}
\caption{\label{tab:timeml:res} Results on predicting TimeML
  event-event relationships; comparison between word-based baseline
  and disjunctive ensemble models.}
\end{center}
\end{table}

\paragraph{Results} Our results are summarized in
Table~\ref{tab:timeml:res}.  We compare the performance of the
disjunctive ensemble from Section~\ref{sec:exper-1:-conf} against a
naive word-based model. Both these models were trained on main and
subordinate clauses from the {\sc Bllip} corpus.  We also report the
accuracy of a majority baseline which defaults to the most frequent
class in the {\sc Bllip} training data (i.e.,~{\sc before}). Finally,
we report the performance of a (disjunctive) ensemble model that has
been trained and tested on the TimeBank corpus (see the column
TestInst in Table~\ref{tab:mapping}) using leave-one-out
crossvalidation. Comparison between the latter model and the {\sc
  Bllip}-trained ensemble will indicate whether unannotated data is
indeed useful in reducing annotation effort and training requirements
for temporal interpretation models.

As can be seen, the disjunctive model trained on the {\sc Bllip}
corpus significantly outperforms the two baseline models. It also
outperforms the ensemble model trained on TimeBank by a wide
margin.\footnote{Unfortunately, we cannot use a $\chi^2$ test to
  assess whether the differences between the two ensembles are
  statistically significant due to the leave-one-out crossvalidation
  methodology employed when training and testing on the TimeBank
  corpus. This was necessary given the small size of the event-event
  relation data extracted from TimeBank (2,533 instances in total, see
  Table~\ref{tab:mapping}).} We find these results encouraging
considering the approximations in our temporal interpretation model
and the noise inherent in the {\sc Bllip} training data.  Also note
that, despite being linguistically informed, our feature space encodes
very basic semantic and temporal distinctions.  For example, aspectual
information is not taken into account, and temporal expressions are
not analyzed in detail. One would hope that more extensive feature
engineering would result in improved results.

\begin{table}[t]
\begin{center}
\begin{tabular}{|l|c|d|d|d|} \hline
                      & \multicolumn{2}{c|}{{\sc Bllip}} & \multicolumn{2}{c|}{TimeBank}\\
 TimeMLRel            & Accuracy  & \labe{F-score} & \labe{Accuracy} & \labe{F-score} \\  \hline
 {\sc before}         & 46.4         &   47.6 & 63.2 & 53.2\\
 {\sc begins}         & 10.5         &   7.8  & 0.0  & 0.0   \\
 {\sc ends}           & 14.1         &   3.7  & 4.7  & 7.7\\
 {\sc includes}       & 50.0         &   51.5 & 8.5  & 9.8\\
 {\sc simultaneous}   & 46.7         &   47.8 & 6.7  & 8.9\\
 {\sc no-temp-rel}    & 62.8         &   66.1 & 49.6 & 53.5\\ \hline
      All             & 53.0         &   45.8 & 42.7 & 40.5\\ \hline
\end{tabular}
\caption{\label{tab:per:class} Ensemble results on inferring individual temporal
 relations; comparison between ensemble model trained on {\sc Bllip}
 and TimeBank corpora.}
\end{center}
\end{table}

We further examined how performance varies for each class.
Table~\ref{tab:per:class} provides a comparison between the two
ensemble models trained on {\sc Bllip} and the TimeBank corpus,
respectively. Both models have difficulty with {\sc begins} and {\sc
  ends} classes. This is not entirely surprising, since these classes
are represented by a relatively small number of training instances
(see Table~\ref{tab:mapping}). The two models yield comparable results
for {\sc before}, whereas the {\sc Bllip}-trained ensemble delivers
better performance for {\sc includes}, {\sc simultaneous}, and {\sc
  no-temp-rel}.

We are not aware of any previous work that attempts to do a similar
task. However, it is worth mentioning \citeA{Boguraev:Ando:05} who
consider the interpretation of event-time temporal relations inter-
and intra-sententially. They report accuracies ranging from~53.1\% and
58.8\%~depending on the intervening distance between the events and
the times in question (performance is better for events and times
occurring close to each other). Interestingly, their interpretation
model exploits unannotated corpora \emph{in conjunction} with TimeML
annotations to increase the amount of labeled data for training. Their
method identifies unannotated instances that are distributionally
similar to the manually annotated corpus.  In contrast, we rely solely
on unannotated data during training while exploiting instances
explicitly marked with temporal information. An interesting future
direction is the combination of such data with TimeML annotations as a
basis for devising improved models (for details, see
Section~\ref{sec:discussion}).

\section{General Discussion}
\label{sec:discussion}


In this paper we proposed a data intensive approach to temporal
inference. We introduced models that learn temporal relations from
sentences where temporal information is made explicit via temporal
markers and assessed their potential in inferring relations in cases
where overt temporal markers are absent. Previous work has focused on
the automatic tagging of temporal expressions \cite{Wilson:ea:01}, on
learning the ordering of events from manually annotated data
\cite{Mani:ea:03}, and inferring the temporal relations between events
and time expressions from both annotated and unannotated data
\cite{Boguraev:Ando:05}.

Our models bypass the need for manual annotation by training
exclusively on instances of temporal relations that are made explicit
by the presence of temporal markers.  We compared and contrasted
several models varying in their linguistic assumptions and employed
feature space. We also explored the tradeoff between model complexity
and data requirements. Our results indicate that less sophisticated
models (e.g.,~the disjunctive model) tend to perform reasonably when
utilizing expressive features and training data sets that are
relatively modest in size.  We experimented with a variety of
linguistically motivated features ranging from verbs and their
semantic classes to temporal signatures and argument structure.  Many
of these features were inspired by symbolic theories of temporal
interpretation, which often exploit semantic representations (e.g.,~of
the two clauses) as well as complex inferences over world knowledge
(e.g.,~\citeR{hobbs:etal:1993,lascarides:asher:1993a,kehler:2002}).

Our best model achieved an F-score of~69.1\% on inferring temporal
relations when trained and tested on the {\sc Bllip} corpus in the
context of a pseudo-disambiguation task. This performance is a
significant improvement over the baseline and compares favorably with
human performance on the same task. Detailed exploration of the
feature space further revealed that not only lexical but also
syntactic information is important for temporal inference. This result
is in agreement with \citeA{Soricut:Marcu:03} who find that syntax
trees encode sufficient information to enable accurate derivation of
discourse relations.

We also evaluated our model's performance on the more realistic task
of predicting temporal relations when these are not explicitly
signaled in text. To this end, we evaluated a {\sc Bllip}-trained
model against TimeBank, a corpus that has been manually annotated with
temporal relations according to the TimeML specifications.  This
experimental set-up was challenging from many perspectives. First,
some of the temporal markers used in our study received multiple
meanings.  The ambiguity unavoidably introduced a certain amount of
noise in estimating the parameters of our model and defining a mapping
between markers and TimeML relations. Second, there is no guarantee
that the relations signaled by temporal markers connecting main and
subordinate clauses hold for events attested in other syntactic
configurations such as non-temporal subordination or coordination.
Given these approximations, our model performed reasonably, reaching
an overall F-score of~45.8\% on the temporal inference task and
showing best performance for relations {\sc before}, {\sc includes},
{\sc simultaneous} and {\sc no-temp-rel}. These results show that it
is possible to infer temporal information from corpora even if they
are not semantically annotated in any way and hold promise for
relieving the data acquisition bottleneck associated with creating
temporal annotations.

An important future direction lies in modeling the temporal relations
of events across sentences. In order to achieve full-scale temporal
reasoning, the current model must be extended in a number of ways.
These involve the incorporation of extra-sentential information to the
modeling task as well as richer temporal information (e.g.,~tagged
time expressions; see \citeR{Mani:ea:03}). The current models perform
the inference task independently of their surrounding context.
Experiment~\ref{exp:human} revealed this is a rather difficult task;
even humans cannot easily make decisions regarding temporal relations
out-of-context. In future work, we plan to take into account
contextual (lexical and syntactic) as well as discourse-based features
(e.g.,~coreference resolution).  Many linguists have also observed
that identifying the {\em discourse structure} of a text,
conceptualized as a hierarchical structure of rhetorically connected
segments, and identifying the temporal relations among its events are
logically co-dependent tasks
\cite<e.g.,>{kamp:reyle:1993,hobbs:etal:1993,lascarides:asher:1993a}.
For example, the fact that we interpret (\ref{ex:temp}a) as forming a
{\em narrative} with (\ref{ex:temp}c) and (\ref{ex:temp}c) as
providing {\em background} information to (\ref{ex:temp}b) yields the
temporal relations among the events that we described in
Section~\ref{sec:introduction}: namely, the temporal progression
between kissing the girl and walking home, and the temporal overlap
between remembering talking to her and walking home.

\ex{
\item [\ref{ex:temp}]
\ex{
\item John kissed the girl he met at a party. 
\item Leaving the party, John walked home.
\item He remembered talking to her and asking her for her name. 
}}

This logical relationship between discourse structure and temporal
structure suggests that the output of a {\em discourse parser}
\cite<e.g.,>{marcu:1999,Soricut:Marcu:03,baldridge:lascarides:2005a}
could be used as an informative source of features for inferring
temporal relations across sentence boundaries.  This would be
analogous at the discourse level to the use we made here of a
sentential parser as a source of features in our experiments for
inferring sentence-internal temporal relations.

The approach presented in this paper can also be combined with the
annotations present in the TimeML corpus in a semi-supervised setting
similar to \citeA{Boguraev:Ando:05} to yield improved performance.
Another interesting direction for future work would be to use the
models proposed here in a bootstrapping approach. Initially, a model
is learned from unannotated data and its output is manually edited
following the ``annotate automatically, correct manually'' methodology
used to provide high volume annotation in the Penn Treebank project. At
each iteration the model is retrained on progressively more accurate
and representative data. Another issue related to the nature of our
training data concerns the temporal information entailed by some of
our markers which can be ambiguous.  This could be remedied either
heuristically as discussed in Section~\ref{sec:data-extraction} or by
using models trained on unambiguous markers (e.g.,~\word{before},
\word{after}) to disambiguate instances with multiple readings.
Another possibility is to apply a separate disambiguation procedure on
the training data (i.e.,~prior to the learning of temporal inference
models).

Finally, we would like to investigate the utility of these temporal
inference models within the context of specific natural language
processing applications.  We thus intend to explore their potential in
improving the performance of a multi-document summarisation system.
For example, a temporal reasoning component could be useful not only
for extracting temporally congruent events, but also for structuring
the output summaries, i.e.,~by temporally ordering the extracted
sentences. Although the models presented here target primarily
interpretation tasks, they could also be adapted for generation tasks,
e.g.,~for inferring if a temporal marker should be generated and where
it should be placed.



\section*{Acknowledgments}

This work was supported by {\sc EPSRC} (Lapata, grant~{\sc
  GR/T04540/01}; Lascarides, grant~{\sc GR/R40036/01}).  We are
grateful to Regina Barzilay and Frank Keller for helpful comments and
suggestions. Thanks to the anonymous referees whose feedback helped to
substantially improve the present paper.  A preliminary version of
this work was published in the proceedings of NAACL 2004; we also
thank the anonymous reviewers of that paper for their comments.

\newpage
\section*{Appendix A. Experimental Materials for Human Evaluation}

The following is the list of materials used in the human evaluation
study reported in Experiment~\ref{exp:human}
(Section~\ref{sec:experiment-2:-human}). The sentences were extracted
from the {\sc Bllip} corpus following the procedure described in
Section~\ref{sec:data-extraction}.

\begin{table}[h!]
\begin{center}
\begin{small}
\begin{tabular}{|lp{14.3cm}|} \hline
1 & In addition, agencies weren't always efficient in getting the word
to other agencies \raisebox{-.1cm}{\parbox{.8cm}{\hrulefill}} the
company was barred. \hfill {\bf when}\\ 
2 & Mr. Reagan learned of the news \raisebox{-.1cm}{\parbox{.8cm}{\hrulefill}} National Security Adviser Frank Carlucci called to tell him he'd seen it on television. \raisebox{-.1cm}{\parbox{.8cm}{\hrulefill}} {\bf when} \\
3 & For instance, National Geographic caused an uproar \raisebox{-.1cm}{\parbox{.8cm}{\hrulefill}} it used a computer to neatly move two Egyptian pyramids closer together in a photo. \hfill {\bf when} \\
4 & Rowes Wharf looks its best \raisebox{-.1cm}{\parbox{.8cm}{\hrulefill}} seen from the new Airport Water Shuttle
 speeding across Boston harbor. \hfill {\bf when} \\
5 & More and more older women are divorcing
\raisebox{-.1cm}{\parbox{.8cm}{\hrulefill}} their husbands
retire. \hfill {\bf when}\\ \hline \hline
6& Together they prepared to head up a Fortune company \raisebox{-.1cm}{\parbox{.8cm}{\hrulefill}} enjoying a tranquil country life. \hfill {\bf while}\\
7& \raisebox{-.1cm}{\parbox{.8cm}{\hrulefill}} it has been estimated that 190,000 legal abortions to adolescents occurred, an unknown number of illegal and unreported abortions took place as well. \hfill {\bf while}\\
8& Mr. Rough, who is in his late 40s, allegedly leaked the information \raisebox{-.1cm}{\parbox{.8cm}{\hrulefill}} he served as a New York Federal Reserve Bank director from January 1982 through December 1984. \hfill {\bf while}\\
9& The contest became an obsession for Fumio Hirai, a 30-year-old mechanical engineer, whose wife took to ignoring him \raisebox{-.1cm}{\parbox{.8cm}{\hrulefill}} he and two other men tinkered for months with his dancing house plants.  \hfill {\bf while}\\
10& He calls the whole experience ``wonderful, enlightening,
fulfilling'' and is proud that MCI functioned so well
\raisebox{-.1cm}{\parbox{.8cm}{\hrulefill}} he was gone.  \hfill {\bf
  while}\\ \hline \hline
11& A lot of them want to get out \raisebox{-.1cm}{\parbox{.8cm}{\hrulefill}} they get kicked out.  \hfill {\bf before}\\
12& \raisebox{-.1cm}{\parbox{.8cm}{\hrulefill}} prices started falling, the market was doing \$1.5 billion a week in new issues, says the head of investment banking at a major Wall Street firm.  \hfill {\bf before}\\
13& But \raisebox{-.1cm}{\parbox{.8cm}{\hrulefill}} you start feeling sorry for the fair sex, note that these are the Bundys, not the Bunkers.   \hfill ~~~~{\bf before}\\
14& The Organization of Petroleum Exporting Countries will travel a rocky road \raisebox{-.1cm}{\parbox{.8cm}{\hrulefill}} its Persian Gulf members again rule world oil markets.  \hfill {\bf before}\\
15& Are more certified deaths required
\raisebox{-.1cm}{\parbox{.8cm}{\hrulefill}} the FDA acts? \hfill {\bf
  before}\\ \hline \hline
16& Currently, a large store can be built only \raisebox{-.1cm}{\parbox{.8cm}{\hrulefill}} smaller merchants in the area approve it, a difficult and time consuming process.  \hfill {\bf after}\\
17& The review began last week
\raisebox{-.1cm}{\parbox{.8cm}{\hrulefill}} Robert L. Starer was named
president. \hfill {\bf after}\\
18& The lower rate came \raisebox{-.1cm}{\parbox{.8cm}{\hrulefill}} the nation's central bank, the Bank of Canada, cut its weekly bank rate to 7.2\% from 7.54\%.  \hfill {\bf after}\\
19& Black residents of Washington's low-income Anacostia section forced a three-month closing of a Chinese-owned restaurant \raisebox{-.1cm}{\parbox{.8cm}{\hrulefill}} the owner threatened an elderly black woman customer with a pistol.  \hfill {\bf after}\\
20& Laurie Massa's back hurt for months
\raisebox{-.1cm}{\parbox{.8cm}{\hrulefill}} a delivery truck slammed
into her car in 1986.  \hfill {\bf after}\\ \hline
\end{tabular}
\end{small}
\caption{\label{tab:interprete:1} Materials for the temporal
  pseudo-disambiguation task; markers in bodlface indicate the gold
  standard completion; subjects were asked to select the missing word
  from the set of temporal markers \{\word{after, before, while, when,
    as, once, until, since}\}}
\end{center}
\end{table}

\addtocounter{table}{-1}
\begin{table}
\begin{center}
\begin{small}
\begin{tabular}{|lp{14.2cm}|} \hline
21& Donald Lasater, 62, chairman and chief executive office, will assume the posts Mr. Farrell vacates \raisebox{-.1cm}{\parbox{.8cm}{\hrulefill}} a successor is found.  \hfill {\bf until}\\
22& The council said that the national assembly will be replaced with appointed legislators and that no new elections will be held \raisebox{-.1cm}{\parbox{.8cm}{\hrulefill}} the U.S. lifts economic sanctions. \hfill {\bf until}\\
23& \raisebox{-.1cm}{\parbox{.8cm}{\hrulefill}} those problems disappear, Mr. Melzer suggests working with the base, the raw material for all forms of the money supply. \hfill {\bf until}\\
24& A green-coffee importer said there is sufficient supply in Brazil \raisebox{-.1cm}{\parbox{.8cm}{\hrulefill}} the harvest gets into full swing next month. \hfill {\bf until}\\
25& They will pump \raisebox{-.1cm}{\parbox{.8cm}{\hrulefill}} the
fire at hand is out. \hfill {\bf until}\\ \hline \hline
26& \raisebox{-.1cm}{\parbox{.8cm}{\hrulefill}} the gene is inserted
in the human TIL cells, another safety check would be made. \hfill
{\bf once} \\
27&  \raisebox{-.1cm}{\parbox{.8cm}{\hrulefill}} part of a bus system is subject to market discipline, the entire operation tends to respond. \hfill {\bf once}\\
28& In China by contrast, \raisebox{-.1cm}{\parbox{.8cm}{\hrulefill}} joint ventures were legal, hundreds were created. \hfill {\bf once}\\
29 & The company said the problem goes away \raisebox{-.1cm}{\parbox{.8cm}{\hrulefill}} the car warms up. \hfill {\bf once}\\
30& \raisebox{-.1cm}{\parbox{.8cm}{\hrulefill}} the Toronto merger is complete, the combined entity will have 352 lawyers. \hfill {\bf once}\\\hline\hline
31& The justices ruled that his admission could be used \raisebox{-.1cm}{\parbox{.8cm}{\hrulefill}} he clearly had chosen speech over silence. \hfill {\bf since}\\
32& Milosevic's popularity has risen \raisebox{-.1cm}{\parbox{.8cm}{\hrulefill}} he became party chief in Serbia, Yugoslavia's biggest republic, in 1986. \hfill {\bf since}\\
33& The government says it has already eliminated 600 million hours of paperwork a year \raisebox{-.1cm}{\parbox{.8cm}{\hrulefill}} Congress passed the Paperwork Reduction Act in 1980. \hfill {\bf since}\\
34& It was the most serious rebellion in the Conservative ranks \raisebox{-.1cm}{\parbox{.8cm}{\hrulefill}} Mr. Mulroney was elected four years ago. \hfill {\bf since}\\
35& There have been at least eight settlement attempts \raisebox{-.1cm}{\parbox{.8cm}{\hrulefill}} a Texas court handed down its multi-billion dollar judgment two years ago. \hfill {\bf since}\\\hline\hline
36& Brud LeTourneau, a Seattle management consultant and Merit smoker, laughs at himself \raisebox{-.1cm}{\parbox{.8cm}{\hrulefill}} he keeps trying to flick non-existent ashes into an ashtray. \hfill {\bf as}\\
37& Britain's airports were disrupted \raisebox{-.1cm}{\parbox{.8cm}{\hrulefill}} a 24-hour strike by air traffic control assistants resulted in the cancellation of more thank 500 flights and lengthy delays for travelers. \hfill {\bf as}\\
38& Stocks plunged \raisebox{-.1cm}{\parbox{.8cm}{\hrulefill}} investors ignored cuts in European interest rates and dollar and bond rallies. \hfill {\bf as}\\
39& At Boston's Logan Airport, a Delta plane landed on the wrong runway \raisebox{-.1cm}{\parbox{.8cm}{\hrulefill}} another jet was taking off. \hfill {\bf as}\\
40& Polish strikers shut Gdansk's port
\raisebox{-.1cm}{\parbox{.8cm}{\hrulefill}} Warsaw rushed riot police
to the city. \hfill {\bf as} \\\hline
\end{tabular}
\end{small}
\caption{\label{tab:interprete:2} ({\em continued})}
\end{center}
\end{table}


\bibliography{references}

\bibliographystyle{theapa}
\end{document}